\documentclass{article}
\usepackage{iclr2027_conference,times}

\usepackage{amsmath,amsfonts,bm}

\def\eqref#1{equation~\ref{#1}}

\def\1{\bm{1}}

\DeclareMathAlphabet{\mathsfit}{\encodingdefault}{\sfdefault}{m}{sl}
\SetMathAlphabet{\mathsfit}{bold}{\encodingdefault}{\sfdefault}{bx}{n}

\usepackage{enumitem}
\usepackage{hyperref}
\usepackage{url}
\usepackage[most]{tcolorbox}
\usepackage{tikz}
\usetikzlibrary{arrows.meta, positioning, fit, backgrounds}
\usepackage[most]{tcolorbox}
\usepackage{amsmath, amssymb}
\usepackage{booktabs}
\usepackage{graphicx}
\usepackage{algorithm}
\usepackage{algorithmic}
\usepackage{pgfplots}
\pgfplotsset{compat=1.18}

\usepackage{amsthm}
\newtheorem{proposition}{Proposition}
\newtheorem{corollary}{Corollary}

\usepackage{xcolor}

\title{Learning Compact State Abstractions for LLM Agents}

\author{Yaswanth Chittepu$^{1}$\thanks{Work done during an internship at Adobe Research. Correspondence to \texttt{ychittepu@umass.edu}.}, \
Raghavendra Addanki$^{2}$, Anup Rao$^{2}$, Branislav Kveton$^{2}$, Scott Niekum$^{1}$ \\
$^{1}$University of Massachusetts Amherst \quad $^{2}$Adobe Research
}

\usepackage[most]{tcolorbox}
\usepackage{listings}
\usepackage{xcolor}

\lstdefinestyle{phicode}{
    language=Python,
    basicstyle=\ttfamily\scriptsize,
    breaklines=true,
    columns=fullflexible,
    keepspaces=true,
    showstringspaces=false,
    aboveskip=0pt,
    belowskip=0pt
}

\newtcblisting{phibox}[1]{
  breakable,
  title={#1},
  listing only,
  colback=gray!5,
  colframe=black,
  boxrule=0.5pt,
  arc=2pt,
  left=4pt,
  right=4pt,
  top=4pt,
  bottom=4pt,
}

\iclrfinalcopy
\begin{document}
\raggedbottom

\maketitle
\lhead{Preprint}

\begin{abstract}
LLM agents commonly use the full interaction history as a proxy for state, retaining all past observations and actions regardless of whether they remain relevant. We instead ask what information an agent actually needs to carry forward. We define agent state as a compact abstraction of interaction history that preserves the competence of a fixed reference agent, with the minimal such abstraction as the ideal. Because matching the reference agent's behavior is unnecessarily restrictive and difficult for black-box policies, we define sufficiency through task performance rather than behavior matching. We represent state abstractions as deterministic recursive programs and search over them using an LLM-guided genetic algorithm. Across ALFWorld and ScienceWorld, the learned abstractions preserve near-reference task success while reducing mean policy input by up to 71\% and 57\%, respectively. The resulting programs are directly inspectable, providing an explicit account of what information is retained, updated, and discarded during interaction.
\end{abstract}

\section{Introduction}

LLM agents typically select an action by conditioning on the full interaction
history. Observations, actions, and intermediate reasoning are appended to a
growing context and passed back to the model at every step \citep{yao2023react}.
This is a natural default because the history is sufficient by construction: no
information from the interaction is intentionally discarded. But it retains all
information indiscriminately, regardless of whether it remains relevant. As
context grows, it accumulates observations and action outcomes whose long-term
value may be much smaller than the space they occupy. The model must repeatedly
recover the useful information from an increasingly large history, which can
eventually exceed the model's finite context window. Most existing approaches
treat this as a context-management problem. They summarize the history, retrieve
selected memories, or compress the context once it becomes too large
\citep{jiang2023llmlingua,wu2021recursively,kang2025acon}.

We take a different view. The more fundamental question is not how to shorten the context, but what information the agent actually needs to carry forward. The notion of state is foundational to sequential decision making, from
Markov decision processes and partially observable MDPs
\citep{bellman57dynamic,sondik71thesis} to reinforcement learning
\citep{sutton88learning,watkins92qlearning,williams92simple}. Sequential decision making provides a natural framing for this question through state abstraction \citep{li2006towards}: rather than assuming the full interaction history is the agent's state, we ask whether a smaller representation can preserve the competence of the same agent on the task.

A natural answer is to define agent state in terms of what the agent needs to remain competent. We call a compact abstraction of the interaction history an agent state if the reference agent preserves its performance when conditioned on that abstraction rather than the full history; the ideal agent state is the minimal such abstraction. This separates two requirements. Sufficiency asks whether the abstraction preserves competence, while minimality asks how much information can be removed without sacrificing it. Because many tasks admit multiple successful trajectories, we measure sufficiency through preserved task performance rather than exact behavioral matching. The full history trivially satisfies sufficiency, so the problem becomes meaningful only when both sufficiency and minimality are considered together. We instantiate the abstraction as a deterministic recursive program that updates the state throughout the interaction and search over such programs with a genetic algorithm, yielding explicit state-update rules that can be inspected and executed without additional model calls at deployment.

Our contributions are threefold:
\begin{itemize}[
    leftmargin=1em,
    labelsep=0.4em,
    topsep=2pt,
    itemsep=1pt,
    parsep=0pt
]
\item \textbf{Agent state abstraction.} We formulate agent state for black-box LLM agents as a compact abstraction of interaction history that preserves the competence of a fixed reference agent. This connects the agentic setting to classical state abstraction \citep{li2006towards}, with sufficiency defined through task performance.
\item \textbf{Programmatic state discovery.} We represent agent states as deterministic recursive programs and develop an LLM-guided genetic search procedure \citep{romeraparedes2024funsearch, guo2024evoprompt} that discovers such abstractions from interaction alone.

\item \textbf{Empirical evaluation and inspectability.}
We evaluate the approach on ALFWorld \citep{shridhar2021alfworld} and
ScienceWorld \citep{wang2022scienceworld}. Across ten task families, the learned
abstractions reduce mean policy input by up to 71\% in ALFWorld and 57\% in
ScienceWorld. 
Averaged over three evaluation runs, we detect no systematic difference in success from full history ($+1.7$ pp, 95\% CI $[-1.8, +5.3]$), and the agent using the learned state abstraction has the lowest run-to-run variance of the evaluated methods (Appendix~\ref{app:success_uncertainty}).
Because each state is
an executable program, it also provides an inspectable description of the
information carried forward during interaction.

\end{itemize}

\section{Background and Related Work}
\label{sec:background}

\subsection{LLM Agents and the Context-as-State Convention}

LLM agents typically choose each action by conditioning on the accumulated interaction history. ReAct \citep{yao2023react} interleaves reasoning traces with actions and appends both to a growing context, and subsequent agent scaffolds, including reflection over failed attempts \citep{shinn2023reflexion}, planner--executor decompositions \citep{prasad2024adapt}, and skill libraries \citep{wang2023voyager}, elaborate the control flow while retaining this convention. In effect, the context serves as the agent's state.

Retaining the full history is sufficient by construction, since no information from the interaction is intentionally discarded. It is also maximally non-selective. As an episode progresses, the context accumulates observations and action outcomes whose durable information may be much smaller than their textual representation, forcing the model to repeatedly recover the relevant information from an increasingly large history.

Existing agent-memory systems impose structure on this history through external storage, retrieval, or hand-designed memory components. MemGPT \citep{packer2023memgpt} manages a bounded working context against external storage, while generative agents maintain a retrievable memory stream with recency and relevance scoring \citep{park2023generative}. These approaches improve how information is stored or accessed, but the structure and contents of the state are still largely specified by the system designer. Other work learns or prompts explicit memory and state for agents. MCMA \citep{liang2026learningremembermetacognitivemanagement} trains a memory copilot to abstract past trajectories into reusable cross-episode procedural memory. StateAct \citep{rozanov2025stateactenhancingllmbase} prompts the acting model to describe its own goal and state at every step, following hand-annotated few-shot examples; this self-reported state is neither learned nor tested for sufficiency as a replacement for the interaction history. Our focus is instead on identifying what information the agent actually needs to carry forward.

\subsection{State Abstraction in Sequential Decision Making}

A central question in sequential decision making is what information from the past must be retained for future control. In partially observable environments, the full interaction history is always a valid information state, but it grows with the horizon and may contain substantial redundancy. Classical work therefore seeks smaller representations that preserve only the distinctions that matter for decision making.

One line of work formalizes this through state abstraction. \citet{li2006towards} distinguish several abstraction schemes according to what they preserve. Model-irrelevance abstractions preserve immediate rewards and transition dynamics over abstract states. $Q$-irrelevance abstractions preserve action values, while $\pi$-irrelevance abstractions preserve the action selected by a policy. These form increasingly weaker notions of equivalence, allowing progressively coarser abstractions as fewer properties of the original state space must be retained.

Bisimulation provides a closely related notion of state equivalence. \citet{ferns2012metrics} characterize states as similar when they induce similar immediate rewards and transition behavior over equivalent future states, and extend this binary relation to a quantitative metric. In this sense, bisimulation is closely aligned with model-preserving abstractions, but provides a recursive notion of behavioral similarity between states. \citet{abel2016near} further study approximate abstractions, showing how violations of exact abstraction conditions translate into bounded degradation in control performance.

Policy-relative notions are especially relevant to our setting. \citet{castro2020scalable} introduces a policy-dependent variant of bisimulation in which state similarity is defined relative to a fixed behavior policy. This weakens the requirement that the abstraction preserve distinctions that are irrelevant under the particular policy being executed. Our setting is similarly policy-relative: the LLM agent is fixed, and we seek an abstraction sufficient for that specific agent rather than for all possible policies. However, unlike policy-relative bisimulation, we do not assume access to transition dynamics or state-level reward structure, and we do not require the abstracted process to preserve the environment model.

A related line of work studies compact representations of history under partial observability. \citet{littman2001predictive} represent state through predictions of future observable events rather than latent environment variables, while \citet{singh2003learning} develop methods for learning such predictive summaries from experience. \citet{subramanian2022approximate} study recursively updateable summaries of history that approximately preserve the information required for planning and control. Our state representation shares this recursive structure, but our sufficiency criterion is defined relative to the realized performance of a fixed black-box agent rather than preservation of model-based planning quantities.

These approaches differ in what an abstraction is required to preserve, including transition dynamics, rewards, values, policies, or predictive distributions. Our setting differs in both what is available and what is fixed. The environment and LLM policy are treated as black boxes, the policy is not relearned after abstraction, and the abstraction itself is the object of optimization. We therefore seek a policy-relative notion of sufficiency that can be evaluated directly from interaction, based on whether the fixed agent preserves its task performance under the abstract state.

\subsection{Context Compression for Agents}

A parallel line of work reduces the amount of context presented to language models. Prompt-compression methods such as LLMLingua \citep{jiang2023llmlingua} remove low-importance tokens under a compression budget, while learned compressors such as \citet{chevalier2023adapting} encode long contexts into compact summary vectors that are provided to the model as soft prompts. Recursive summarization methods \citep{wu2021recursively} have also been used to process inputs beyond the model's context window by hierarchically summarizing segments and their summaries. More directly related to our setting, ACON \citep{kang2025acon} learns natural-language compression guidelines from successful and failed agent trajectories and distills the resulting compressor into a smaller model.

These approaches primarily treat history reduction as a compression or memory-management problem: given a context budget or compression mechanism, the goal is to retain enough information for the agent to continue performing well. Our perspective is different. We treat the problem as one of state identification: rather than asking how much of the history can be compressed, we ask what information must be carried forward for a fixed agent to preserve its competence. Context reduction is therefore a consequence of identifying a sufficient state, rather than the primary objective.

A second distinction is when abstraction is applied. Many context-compression methods operate only after the accumulated history reaches a budget or threshold, compressing an already-grown context. In our setting, the agent operates on the abstract state from the first step onward: each new observation is incorporated into the recursively maintained state, and the discarded history is never carried forward. The agent therefore acts throughout the episode in the learned state space rather than switching from full-history conditioning to compression once the context becomes large. Because the state update is a deterministic program, this recurrence also requires no additional language-model inference at deployment.

\subsection{Program Search with Language Models}

Recent work has used language models as proposal operators for search over discrete objects. FunSearch \citep{romeraparedes2024funsearch} evolves executable programs by repeatedly generating candidate modifications and evaluating them with an external objective. EvoPrompt \citep{guo2024evoprompt} similarly combines evolutionary search with LLM-generated mutations and recombinations, but over prompt space rather than programs. Other methods, including TextGrad \citep{yuksekgonul2024textgrad} and GEPA \citep{agrawal2026gepa}, use natural-language feedback to iteratively revise prompts or other textual artifacts based on execution outcomes. ALMA \citep{xiong2026learningcontinuallylearnmetalearning} applies LLM-guided search to memory designs for continual learning, selecting programs that store and retrieve experience across a stream of tasks. Our search instead targets a policy-relative, within-episode state that can replace the raw interaction history while preserving a fixed agent's task success.

Our optimization procedure follows this general pattern but applies it to a different object: a recursive state-update program embedded inside an agent-environment interaction loop. Candidate abstractions are evaluated through the downstream return of the fixed agent, and LLM-based mutation, crossover, and reflection are used to search over programs that preserve performance while reducing state complexity.

\section{Method}

We consider a fixed black-box LLM agent $\pi$ interacting with an environment $\mathcal{E}$. At time step $t$, the agent has received observations $o_{1:t}$ and produced actions $a_{1:t-1}$. Standard LLM agents condition on the full interaction history,
\[
c_t = (o_1, a_1, \ldots, o_{t-1}, a_{t-1}, o_t),
\]
which is sufficient by construction but becomes increasingly redundant as the interaction proceeds.

Rather than treating $c_t$ as the agent's state, we seek a compact state $s_t$ that preserves the agent's task performance. The policy $\pi$ remains fixed throughout; the only learned object is an abstraction function $\phi$ that recursively updates the state as
\[
s_t = \phi(s_{t-1}, o_t, a_{t-1}),
\]
with $s_0 = \varnothing$. The task instruction is provided separately to the agent as part of its prompt and is not stored in the recurrent state. The recursive form is necessary because computing an abstraction directly from $c_t$ would itself require retaining the full history, defeating the purpose of the abstraction.

Our goal is therefore to identify a compact, recursively maintainable representation under which the fixed agent remains competent. We next formalize this objective, then describe the program class used to represent $\phi$ and the search procedure used to optimize it.

\subsection{Learning a Sufficient Compact State}

\begin{tcolorbox}[colback=gray!5, colframe=gray!50, halign=center]
An agent state is a compact abstraction of the interaction history that preserves the performance of a reference agent acting on the full history. An ideal agent state is the minimal such abstraction.
\end{tcolorbox}

A natural notion of sufficiency would require the abstracted agent to reproduce the behavior of the reference agent acting on the full history. In our setting, however, behavioral equivalence is unnecessarily strong. Many tasks admit multiple successful trajectories, so two agents may take different actions while achieving the same outcome. Moreover, matching trajectory distributions is impractical for a black-box language model, since an abstraction can alter the agent's subsequent actions and thereby induce a different state distribution from that of the full-history agent.

We therefore define sufficiency through outcome preservation: an abstraction is
useful if the fixed agent retains its task performance when conditioned on the
abstract state. We write $\pi \circ \phi$ for the induced agent that recursively
maintains
$s_t=\phi(s_{t-1},o_t,a_{t-1})$
and acts according to $a_t\sim\pi(\cdot\mid s_t)$.

Let $J(\pi\circ\phi)$ denote the expected task return of this induced agent, and
let $T(\phi)$ denote its mean policy input token usage. We optimize
\begin{equation}
\begin{aligned}
\max_{\phi}\quad &
J(\pi\circ\phi)-\lambda T(\phi) \\
\text{s.t.}\quad &
J(\pi\circ\phi)\geq(1-\epsilon)J(\pi_{\mathrm{full}}), \\
&
T(\phi) < T(\phi_{\mathrm{naive}}),
\end{aligned}
\label{eq:state_abstraction_objective}
\end{equation}
where $\pi_{\mathrm{full}}$ denotes the same fixed policy conditioned on the
full interaction history and $\phi_{\mathrm{naive}}$ is the initial naive
aggregation function. The success constraint prevents state-size reductions
from being achieved through arbitrary degradation in task performance, while
$\lambda$ provides additional pressure toward compact states among candidate
abstractions.

Optimizing Equation~\ref{eq:state_abstraction_objective} over $\phi$ has the
structure of a reinforcement learning problem: the abstraction is applied
inside the interaction loop, its effect on behavior compounds over time, and
its quality is observed only through trajectory-level return. Unlike standard
reinforcement learning, however, the policy $\pi$ is fixed; the learned object
is the state abstraction $\phi$ itself.

\begingroup

\paragraph{Relation to behavior and performance matching.}
Let $h_t$ denote the full interaction history at step $t$, and let $s_t$ be the state that $\phi$ computes from it. The induced agent $\pi\circ\phi$ acts according to $\pi(\cdot \mid s_t)$, while the full-history agent acts according to $\pi_{\mathrm{full}}(\cdot \mid h_t)$. We compare our criterion with two stronger alternatives. Let $d_t^{\pi\circ\phi}$ denote the distribution of histories $h_t$ induced by $\pi\circ\phi$, and let $\mathrm{TV}$ denote the total variation distance. Behavior matching controls the cumulative mismatch between the two agents' action distributions,
\begin{equation}
B(\phi) = \sum_{t=1}^{H} \mathbb{E}_{h_t \sim d_t^{\pi\circ\phi}} \Big[ \mathrm{TV}\big( \pi_{\mathrm{full}}(\cdot \mid h_t),\, \pi(\cdot \mid s_t) \big) \Big].
\end{equation}
Performance matching instead controls the absolute success gap $|J(\pi_{\mathrm{full}}) - J(\pi\circ\phi)|$.

\begin{proposition}[Hierarchy of sufficiency criteria]\label{prop:hierarchy}
Let task success be a terminal binary reward over horizon $H$. For any abstraction $\phi$,
\begin{equation}
J(\pi_{\mathrm{full}}) - J(\pi\circ\phi) \;\le\; \big| J(\pi_{\mathrm{full}}) - J(\pi\circ\phi) \big| \;\le\; B(\phi).
\label{eq:relaxation_chain}
\end{equation}
For any tolerance $\delta \ge 0$, define
\begin{align*}
\mathcal{F}_{\mathrm{beh}}(\delta) &= \{\phi : B(\phi) \le \delta\}, \\
\mathcal{F}_{\mathrm{perf}}(\delta) &= \{\phi : |J(\pi_{\mathrm{full}}) - J(\pi\circ\phi)| \le \delta\}, \\
\mathcal{F}_{\mathrm{signed}}(\delta) &= \{\phi : J(\pi_{\mathrm{full}}) - J(\pi\circ\phi) \le \delta\}.
\end{align*}
Then $\mathcal{F}_{\mathrm{beh}}(\delta) \subseteq \mathcal{F}_{\mathrm{perf}}(\delta) \subseteq \mathcal{F}_{\mathrm{signed}}(\delta)$.
\end{proposition}

The proof is given in Appendix~\ref{app:proof_hierarchy}. Since $\pi_{\mathrm{full}}$ is a fixed LLM agent, $J(\pi_{\mathrm{full}})$ is a constant, so maximizing the success term $J(\pi\circ\phi)$ in \eqref{eq:state_abstraction_objective} is equivalent to minimizing the signed gap $J(\pi_{\mathrm{full}}) - J(\pi\circ\phi)$, the left-hand side of \eqref{eq:relaxation_chain}. With $\delta = \epsilon J(\pi_{\mathrm{full}})$, $\mathcal{F}_{\mathrm{signed}}(\delta)$ is exactly the feasible set of the success constraint in \eqref{eq:state_abstraction_objective}.

The three criteria differ in what they require an abstraction to preserve. Behavior matching requires the agent's action distributions to remain close throughout the episode. Performance matching allows the behavior to change, but penalizes any change in success, including improvements. Our criterion constrains only the degradation in success. It therefore also admits abstractions that change the agent's behavior while preserving success, as well as abstractions that improve it.

\begin{corollary}[Compactness under relaxation]\label{cor:compactness}
For $c \in \{\mathrm{beh}, \mathrm{perf}, \mathrm{signed}\}$, let $T^*_c = \inf_{\phi \in \mathcal{F}_c(\delta)} T(\phi)$, where $T(\phi)$ is the mean policy input tokens in \eqref{eq:state_abstraction_objective}. Then $T^*_{\mathrm{signed}} \le T^*_{\mathrm{perf}} \le T^*_{\mathrm{beh}}$.
\end{corollary}

\begin{proof}
The feasible sets are nested by Proposition~\ref{prop:hierarchy}, and minimizing the same quantity over a larger feasible set cannot increase its optimal value.
\end{proof}

Equation~\ref{eq:relaxation_chain} therefore gives a sequence of increasingly relaxed sufficiency criteria. Our criterion directly constrains the loss in task success, the quantity our objective optimizes. It has the largest feasible set of the three, and can therefore admit a more compact state than behavior or performance matching.

\endgroup

We view the objective in \eqref{eq:state_abstraction_objective} as a practical surrogate for the ideal of a minimal sufficient state. Our method searches for compact sufficient abstractions within a restricted function class, rather than certifying global minimality.

\subsection{Recursive Program States}

Optimizing Equation~\ref{eq:state_abstraction_objective} requires choosing a function class for $\phi$. We represent $\phi$ as a deterministic Python program that maps the previous state, current observation, and preceding action to the next state,
\begin{equation}
s_t = \phi(s_{t-1}, o_t, a_{t-1}).
\label{eq:recursive_state_update}
\end{equation}

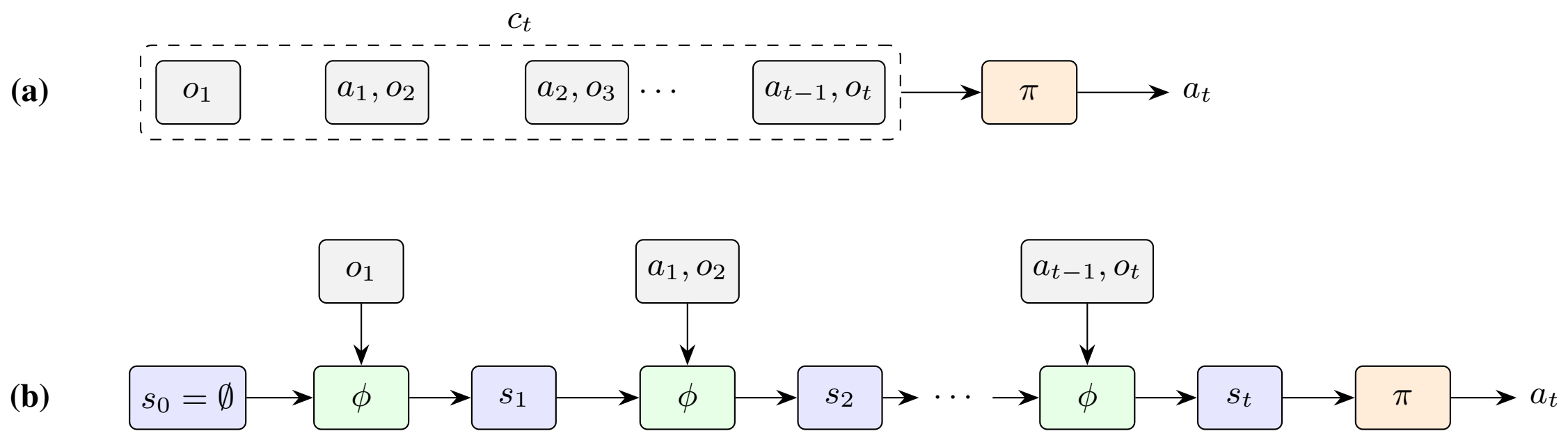
\begin{figure}[t]
\centering
\begin{tikzpicture}[
  font=\small,
  box/.style={draw, rounded corners=2pt, minimum height=6mm, minimum width=8mm, inner sep=3pt},
  obs/.style={box, fill=gray!10},
  st/.style={box, fill=blue!10},
  upd/.style={box, fill=green!10, minimum width=9mm},
  pol/.style={box, fill=orange!15, minimum width=9mm},
  ar/.style={-{Stealth[length=2mm]}, thin},
  >=Stealth
]

\node[font=\small\bfseries] at (-1.6, 1.15) {(a)};
\node[obs] (o1) at (0,   1.15) {$o_{1}$};
\node[obs] (o2) at (1.7, 1.15) {$a_{1}, o_{2}$};
\node[obs] (o3) at (3.6, 1.15) {$a_{2}, o_{3}$};
\node at (4.4, 1.15) {$\cdots$};
\node[obs] (ot) at (5.9, 1.15) {$a_{t-1}, o_t$};

\node[pol] (pi) at (7.9, 1.15) {$\pi$};
\node (at) at (9.5, 1.15) {$a_t$};

\begin{scope}[on background layer]
  \node[draw, dashed, rounded corners=3pt, fit=(o1)(ot), inner sep=4pt] (ctx) {};
\end{scope}
\node[anchor=south, font=\footnotesize] at (ctx.north) {$c_t$};

\draw[ar] (ctx.east) -- (pi.west);
\draw[ar] (pi.east) -- (at.west);

\node[font=\small\bfseries] at (-1.6, -1.75) {(b)};

\node[st]  (s0)   at (-0.1, -1.75) {$s_0=\emptyset$};
\node[upd] (phi1) at (1.55, -1.75) {$\phi$};
\node[st]  (s1)   at (3.0, -1.75) {$s_1$};

\node[upd] (phi2) at (4.65, -1.75) {$\phi$};
\node[st]  (s2)   at (6.1, -1.75) {$s_2$};

\node at (7.2, -1.75) {$\cdots$};

\node[upd] (phit) at (8.45, -1.75) {$\phi$};
\node[st]  (st)   at (9.9, -1.75) {$s_t$};

\node[pol] (pi2)  at (11.45, -1.75) {$\pi$};
\node      (at2)  at (12.8, -1.75) {$a_t$};

\draw[ar] (s0) -- (phi1);
\draw[ar] (phi1) -- (s1);
\draw[ar] (s1) -- (phi2);
\draw[ar] (phi2) -- (s2);
\draw[ar] (s2) -- (6.85, -1.75);
\draw[ar] (7.55, -1.75) -- (phit);
\draw[ar] (phit) -- (st);
\draw[ar] (st) -- (pi2);
\draw[ar] (pi2) -- (at2);

\node[obs] (e1) at (1.55, -0.55) {$o_1$};
\node[obs] (e2) at (4.65, -0.55) {$a_1, o_2$};
\node[obs] (et) at (8.45, -0.55) {$a_{t-1}, o_t$};

\draw[ar] (e1) -- (phi1);
\draw[ar] (e2) -- (phi2);
\draw[ar] (et) -- (phit);

\end{tikzpicture}

\caption{
\textbf{(a)} A standard LLM agent conditions on the full context $c_t$, the
concatenation of all observations and actions, which grows with the interaction.
\textbf{(b)} The abstracted agent maintains a recursive state. At each step,
$\phi$ combines the previous state $s_{t-1}$ with the preceding action
$a_{t-1}$ and current observation $o_t$ to produce $s_t$, which is then provided
to the policy. The raw interaction history is not retained.
}
\label{fig:recursive_state}
\end{figure}

Equation~\ref{eq:recursive_state_update} has the form of a recurrent hidden-state update. At each step, the current observation and preceding action are absorbed into $s_t$, after which the raw interaction history need not be retained. Figure~\ref{fig:recursive_state} contrasts this recursive state representation with the standard full-context agent. Unlike a learned recurrent neural network, however, the update function $\phi$ is an explicit deterministic program whose state variables and update rules are directly inspectable.

Programs are well suited to this setting because the abstraction primarily requires structured state updates: parsing observations, maintaining task-relevant records, updating counters or flags, and discarding text once its relevant information has been extracted. They also allow the abstraction to be executed without an additional model call at each environment step.

We restrict $\phi$ to deterministic programs. Since the abstraction is applied recursively throughout the interaction, stochastic errors introduced at one step could propagate through all subsequent states after the original history has been discarded. Deterministic execution avoids this source of compounding uncertainty.

Finally, the resulting abstraction is directly inspectable. Rather than representing state as an opaque latent vector, the learned program explicitly specifies what information is retained, updated, and discarded during the interaction. This makes the learned abstraction itself a readable hypothesis about the information required by the agent.

\subsection{Searching Over State-Update Programs}
\label{sec:program_search}

The objective in Equation~\ref{eq:state_abstraction_objective} is non-differentiable and can only be evaluated through
interaction with the environment. We therefore optimize over abstraction
programs using a genetic algorithm \citep{holland1975misc-adaptation} with
black-box fitness evaluation, following recent work that combines evolutionary
search with LLM-generated program proposals
\citep{romeraparedes2024funsearch,guo2024evoprompt}.

We maintain a population of candidate state-update programs. Each candidate is
evaluated by rolling out the fixed agent while conditioning on its recursively
maintained state. Parent programs are selected by tournament selection, and
children are generated through mutation or crossover. For mutation, the
proposer, implemented as GPT-5.6 Luna \citep{openai2026gpt56luna}, receives a single parent program; for
crossover, it receives two parents and is asked to combine useful state
components from both. The proposer is additionally given the parent rollouts,
corresponding full-history reference rollouts, and textual feedback from a
reflector model, also implemented as GPT-5.6 Luna. Comparing the abstracted and
reference rollouts helps identify performance-relevant information that may
have been lost by the current abstraction.

\begin{figure}[t]
    \centering
    \includegraphics[width=0.8\linewidth]{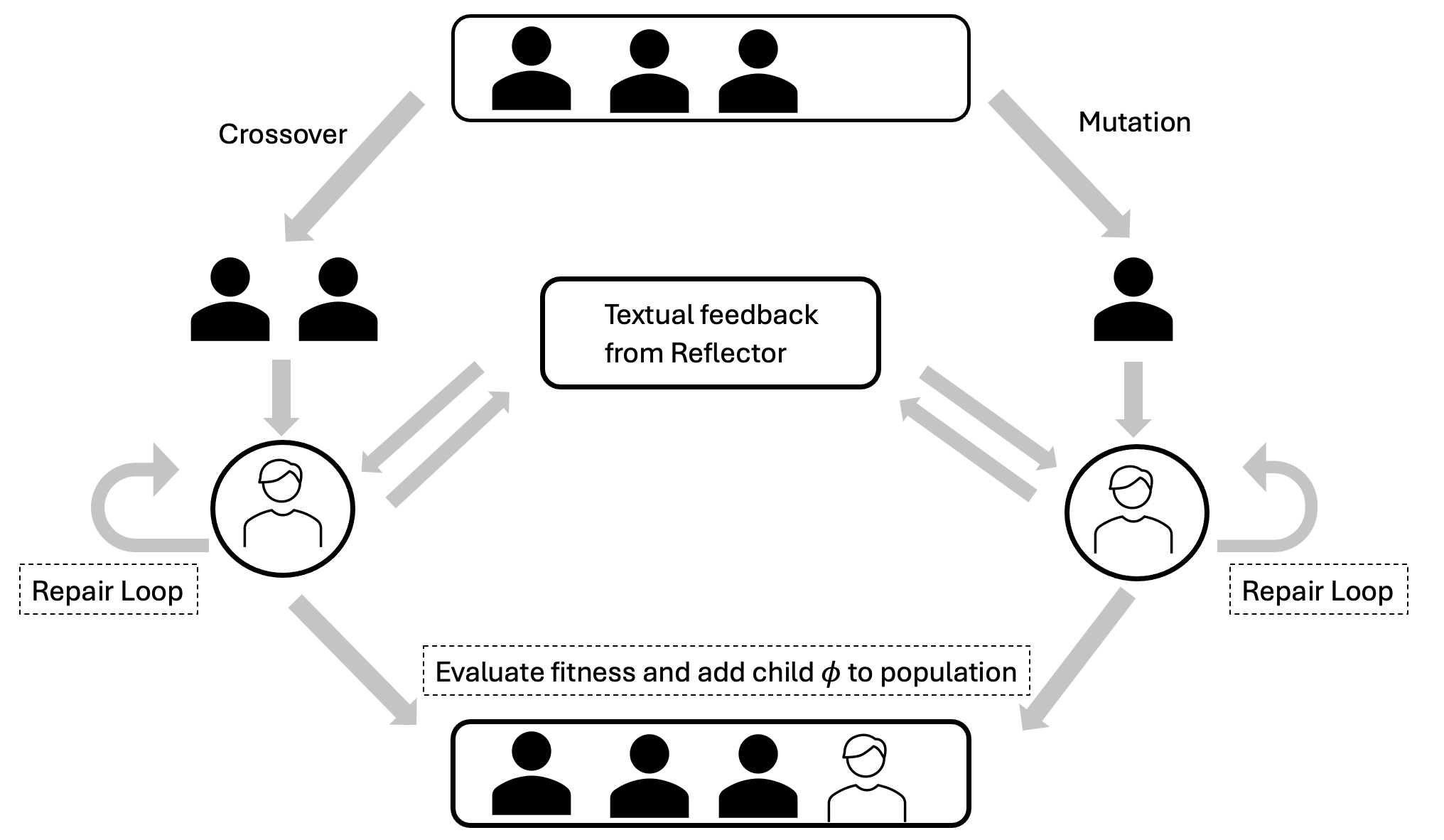}
    \caption{
    Overview of the genetic search procedure. Parent programs are selected for
    mutation or crossover and an LLM proposer generates a child. The child is
    evaluated and, when infeasible, iteratively revised using feedback from a
    reflector for at most three repair attempts. Only the highest-fitness
    candidate from the resulting repair trajectory is added to the population.
    }
    \label{fig:genetic_search}
\end{figure}
\vspace{-1em}
\paragraph{Feasibility.}
We treat preservation of task performance as a constraint rather than allowing
arbitrary trade-offs between success and state size. Let
$J(\pi\circ\phi)$ denote the task success of the fixed policy under abstraction
$\phi$, and let $T(\phi)$ denote its mean policy input token usage. A candidate
is feasible when
\begin{equation}
J(\pi\circ\phi)
    \geq (1-\epsilon)J(\pi_{\mathrm{full}})
\qquad\text{and}\qquad
T(\phi) < T(\phi_{\mathrm{naive}}),
\label{eq:search_feasibility}
\end{equation}
where $\pi_{\mathrm{full}}$ is the full-history reference and
$\phi_{\mathrm{naive}}$ is the initial naive aggregation program. Among
feasible candidates, smaller states are preferred. We use
$\epsilon=0.10$ in all experiments.

\paragraph{Reflection and repair.}
Because small edits to a program can induce large changes in agent behavior,
a newly proposed child is evaluated before being admitted to the population.
If the child is infeasible, the reflector compares its rollout with the
relevant parent and full-history reference rollouts and identifies information
that may have been lost or tracked incorrectly. The proposer then revises the
child using this feedback. We repeat this reflect-and-repair procedure until a
feasible candidate is found or a maximum of three repair attempts is reached.
The original proposal and all repaired variants are retained during this
process, but are not individually inserted into the population. Only the
highest-fitness candidate from the repair process is added.

Each candidate is evaluated only when it is first produced. Its rollout, task
success, token usage, and fitness are cached and reused for subsequent
selection, reflection, mutation, crossover, and repair. After a fixed number
of search iterations, we return the highest-fitness abstraction found during
the search. Search hyperparameters and the reflector and proposer prompts are
provided in Appendix~\ref{app:search_hyperparameters} and \ref{app:prompts} respectively. The first generated program is feasible in only 2 of 10 evaluated families, and all later improvements come from evolutionary proposals (Appendix~\ref{app:search_stats}). Appendix~\ref{app:search_cost} reports the search cost per family. Algorithm~\ref{alg:ga_search} provides the complete search procedure.

\section{Experiments}

\subsection{Setup}
We evaluate whether the learned abstractions preserve agent performance while reducing the number of tokens ingested by the policy across two interactive environments: ALFWorld \citep{shridhar2021alfworld} and ScienceWorld \citep{wang2022scienceworld}. In each domain, we learn a separate abstraction for each predefined task family using training instances from that family and evaluate it on held-out instances. The reference agent uses GPT-5.6 Luna \citep{openai2026gpt56luna} in a ReAct \citep{yao2023react} scaffold and conditions on the full interaction history throughout the episode. We use GPT-5.6 Luna because our goal is to study state abstraction for a fixed
black-box LLM policy rather than compare language models, and Luna provides a
strong agent policy at a tractable cost for the repeated environment rollouts
required by genetic search. Appendix.~\ref{app:cross_model} evaluates the learned abstractions unchanged with a different policy model, Qwen-3.5-397B \citep{qwen2026qwen35}.
\vspace{-1em}
\paragraph{ALFWorld.}
ALFWorld \citep{shridhar2021alfworld} is a text-based household environment in which an agent must complete high-level instructions through navigation and object interaction. We evaluate on its six task families: \texttt{pick\_and\_place\_simple}, \texttt{look\_at\_obj\_in\_light}, \texttt{pick\_clean\_then\_place}, \texttt{pick\_heat\_then\_place}, \texttt{pick\_cool\_then\_place}, and \texttt{pick\_two\_obj\_and\_place}. The benchmark provides held-out environments, allowing us to test whether an abstraction learned for a task family transfers to new instances rather than memorizing individual trajectories.
We use $12$ training instances per task family (drawn from the seen split) and evaluate on all remaining held-out instances---the unused seen environments together with all unseen environments---which range from $19$ to $47$ per family ($47$, $19$, $46$, $27$, $34$, and $29$ for the six families in the order listed above). 
\vspace{-1em}
\paragraph{ScienceWorld.}
ScienceWorld \citep{wang2022scienceworld} contains interactive science tasks requiring an agent to manipulate objects, inspect environments, and execute multi-step procedures. We evaluate on four task families: \texttt{lifespan}, \texttt{forces}, \texttt{classification}, and \texttt{genetics}. These tasks provide a qualitatively different setting from ALFWorld, with state requirements that include experimental progress, object properties, and intermediate scientific observations. We use $24$, $24$, $32$, and $6$ training instances for \texttt{lifespan}, \texttt{forces}, \texttt{classification}, and \texttt{genetics}, respectively (spread-sampled across each family's variation range), and evaluate on $12$, $12$, $16$, and $12$ held-out instances.

\vspace{-1em}
\paragraph{Baselines.}
We compare against the full-history reference agent, an LLM Summarizer baseline,
and ACON \citep{kang2025acon}. The LLM Summarizer uses the same runtime compression mechanism as our
ACON implementation but with a fixed summarization guideline: once the retained
context exceeds a token threshold, an LLM summarizes the older interaction
history while preserving the most recent turn verbatim. ACON additionally
optimizes the summarization guideline using its utility-maximization and
compression-maximization stages before held-out evaluation. Because the original
ACON implementation targets different domains, we faithfully re-implement
its UT$\rightarrow$CO optimization procedure for ALFWorld and ScienceWorld.
\vspace{-1em}
\paragraph{Evaluation.}
We report task success on held-out instances and the mean number of policy input
tokens per step, averaged across evaluation episodes. For the LLM Summarizer
and ACON, this token count reflects only the compressed context provided to the
policy and does not include the additional tokens consumed by the runtime compressor.

\subsection{Results}
\label{sec:results}

Tables~\ref{tab:alfworld_results} and~\ref{tab:scienceworld_results}
compare our learned state abstractions against the full-history reference agent,
a fixed LLM Summarizer, and ACON.

\begin{table}[t]
\centering
\small
\setlength{\tabcolsep}{3.2pt}
\resizebox{\linewidth}{!}{
\begin{tabular}{l cc cc cc cc}
\toprule
& \multicolumn{2}{c}{Full}
& \multicolumn{2}{c}{\textbf{Ours}}
& \multicolumn{2}{c}{LLM-S}
& \multicolumn{2}{c}{ACON} \\
\cmidrule(lr){2-3}
\cmidrule(lr){4-5}
\cmidrule(lr){6-7}
\cmidrule(lr){8-9}
\textbf{ALFWorld task}
& Succ. & Tok.
& Succ. & Tok.
& Succ. & Tok.
& Succ. & Tok. \\
\midrule

Pick \& Place
& \textbf{1.00} & 281
& \underline{0.96} & \textbf{82} (71\%)
& 0.94 & 211 (25\%)
& \textbf{1.00} & \underline{183} (35\%) \\

Look in Light
& \underline{0.95} & 296
& \textbf{1.00} & 247 (17\%)
& \underline{0.95} & \underline{207} (30\%)
& 0.83 & \textbf{200} (32\%) \\

Clean \& Place
& \textbf{0.74} & 519
& \underline{0.72} & \underline{245} (53\%)
& 0.67 & \underline{245} (53\%)
& 0.67 & \textbf{207} (60\%) \\

Heat \& Place
& \underline{0.74} & 555
& \textbf{0.85} & \textbf{172} (69\%)
& 0.63 & 256 (54\%)
& 0.25 & \underline{201} (64\%) \\

Cool \& Place
& \underline{0.76} & 561
& 0.74 & \textbf{167} (70\%)
& 0.59 & 252 (55\%)
& \textbf{0.83} & \underline{216} (61\%) \\

Two Objects
& \textbf{0.90} & 430
& \underline{0.86} & 402 (7\%)
& 0.83 & \underline{245} (43\%)
& 0.42 & \textbf{203} (53\%) \\

\bottomrule
\end{tabular}
}
\caption{
Held-out ALFWorld performance and mean policy input tokens. Numbers in parentheses give the reduction in mean policy input tokens relative to the full-history agent. LLM-S is the fixed LLM Summarizer, and ACON adds learned compression-guideline optimization. Bold marks the best and underline the second-best result for each metric.
}
\label{tab:alfworld_results}
\end{table}

\begin{table}[t]
\centering
\small
\setlength{\tabcolsep}{3.2pt}
\resizebox{\linewidth}{!}{
\begin{tabular}{l cc cc cc cc}
\toprule
& \multicolumn{2}{c}{Full}
& \multicolumn{2}{c}{\textbf{Ours}}
& \multicolumn{2}{c}{LLM-S}
& \multicolumn{2}{c}{ACON} \\
\cmidrule(lr){2-3}
\cmidrule(lr){4-5}
\cmidrule(lr){6-7}
\cmidrule(lr){8-9}
\textbf{ScienceWorld task}
& Succ. & Tok.
& Succ. & Tok.
& Succ. & Tok.
& Succ. & Tok. \\
\midrule

Lifespan
& \textbf{1.00} & 260
& \underline{0.92} & \textbf{136} (48\%)
& \underline{0.92} & 236 (9\%)
& 0.83 & \underline{229} (12\%) \\

Forces
& 0.42 & 394
& \textbf{0.75} & 412 (+5\%)
& 0.33 & \underline{265} (33\%)
& \underline{0.50} & \textbf{208} (47\%) \\

Classification
& \textbf{0.50} & 395
& \underline{0.44} & \textbf{169} (57\%)
& \textbf{0.50} & 246 (38\%)
& 0.38 & \underline{239} (39\%) \\

Genetics
& \underline{0.67} & 401
& \textbf{0.83} & 247 (38\%)
& 0.58 & \underline{237} (41\%)
& 0.58 & \textbf{209} (48\%) \\

\bottomrule
\end{tabular}
}
\caption{
Held-out ScienceWorld performance and mean policy input tokens, with notation as in Table~\ref{tab:alfworld_results}. A $+$ indicates an increase in tokens.
}
\vspace{-1em}
\label{tab:scienceworld_results}
\end{table}

\vspace{-1em}
\paragraph{ALFWorld.}
Our learned abstractions remain close to the full-history reference across all
six task families while substantially reducing policy input for most tasks.
The largest reductions occur for Pick \& Place, Heat \& Place, and Cool \&
Place, where mean policy input falls by approximately 69--71\%. Clean \&
Place is reduced by 53\%, while Look in Light and Two Objects admit much
smaller reductions. This variation suggests that the amount of information
that must be retained is strongly task dependent. The abstractions also generalize beyond the environments used during search. On both seen and unseen environments, the agent using our learned abstraction shows no systematic difference in success from the full-history agent (Appendix.~\ref{app:unseen_envs}).
\vspace{-1em}
\paragraph{ScienceWorld.}
The same pattern appears in ScienceWorld. Our abstractions reduce mean policy
input by 48\% for Lifespan, 57\% for Classification, and 38\% for Genetics,
while maintaining comparable task success. Forces is the exception:
the learned representation is slightly larger than full history. Its success gap over full history reflects run-to-run variability rather than the abstraction, since the full-history agent alone scores between 0.42 and 0.92 across runs (Appendix.~\ref{app:success_uncertainty}). Together with the ALFWorld results, this indicates that our search does not find comparably compact abstractions across all task families.
\vspace{-1em}
\paragraph{Longer horizons.} The full history grows with every step, while the state produced by $\phi$ grows far more slowly. These savings therefore grow with the horizon. On a long-horizon BabyAI~\citep{chevalierboisvert2019babyai} task, $\phi$ cuts mean policy input tokens by 95\% (Appendix.~\ref{app:long_horizon}). Averaged over three evaluation runs, the agent using $\phi$ shows no systematic difference in success from the full-history agent (Appendix.~\ref{app:success_uncertainty}).
The agent using $\phi$ is also the most consistent across runs, with a run-to-run standard deviation of 0.7 pp against 1.0 pp for full history.
\vspace{-1em}
\paragraph{Comparison to compression baselines.}
Relative to ACON, our method attains higher success on four of six ALFWorld
task families and uses fewer policy input tokens on three of six. Relative to
the fixed LLM Summarizer, it attains higher success on all six task families,
uses fewer tokens on three, and ties on one. On ScienceWorld, our method
attains higher success than ACON on all four task families and uses fewer policy
input tokens on Lifespan and Classification. Compared with the fixed LLM
Summarizer, it attains higher success on two tasks, ties on Lifespan, and uses
fewer tokens on Lifespan and Classification. Our method also outperforms a token-matched recent-history baseline by 10.0 percentage points, averaged over three evaluation runs (Appendix.~\ref{app:recent_history}).

Policy input token usage and task success should not be interpreted
independently: unsuccessful trajectories may terminate earlier or make less
progress and therefore consume fewer tokens. In addition, \emph{both compression baselines require online LLM inference to
generate summaries during execution, whereas our learned abstraction is a
deterministic Python program that does not require LLM inference}. 

\subsection{What Do the Learned State Abstractions Contain?}
\label{sec:state_analysis}

To understand what the learned abstractions represent, we inspect the state
variables and update rules of the final programs. Across task families, the
programs typically convert raw interaction text into structured persistent
variables and discard observations once their task-relevant information has
been incorporated. Table~\ref{tab:phi_summary} in
Appendix~\ref{app:phi_examples} summarizes the information retained across all
task families, while Table~\ref{tab:phi_complexity} reports the structural
complexity of each learned program. Additional representative $\phi$ programs
for both ALFWorld and ScienceWorld are provided in
Appendix~\ref{app:phi_examples}. 
During held-out evaluation, the policy explicitly refers to a small core of state fields (Appendix~\ref{app:field_usage}), and hiding seven of the eight field groups significantly lowers success, by up to 36.1 percentage points (Appendix~\ref{app:field_masking}). Appendix~\ref{app:babyai_phi} describes the state learned for the long-horizon BabyAI task.

For example, the ALFWorld Pick \& Place abstraction maintains a structured
search state containing the current location, searched receptacles, target
location, inventory status, and task phase:

\begin{phibox}{ALFWorld Pick \& Place}
m = re.search(r"\byou arrive at\s+(.+?)(?:\.|$)", ol)
if m:
    loc = norm(m.group(1))
    focus = loc
    add_receptacle(loc)
    if held is None:
        phase = "search"

m = re.search(
    r"\b(?:on|in)\s+the\s+([a-z][a-z0-9]*(?:\s+\d+)?)"
    r"\s*,?\s*you see\s+(.*?)(?:\.|$)", ol,
)
if m:
    view = norm(m.group(1))
    mark_searched(view)
    found = matching_target(m.group(2))

    if found and held is None:
        objloc = view
        phase = "pickup"
        frontier = []
\end{phibox}

Other ALFWorld abstractions retain task-specific persistent properties such as
cleanliness, temperature, cooling status, receptacle contents, and task
completion. ScienceWorld yields qualitatively different representations. The
learned states track developmental stages, device configurations, object
relations, entity locations, and experimental progress. For Classification,
for example, the abstraction resolves entities and their locations from raw
observations and maintains a bounded symbolic world model; its learned program
is shown in Appendix~\ref{app:phi_examples}.

In contrast, the abstractions learned for ALFWorld Two Objects and ScienceWorld Forces remain close to bounded recent histories, consistent with their limited token reductions. Overall, the learned programs replace raw trajectory text with task-specific persistent state, while the amount and form of retained information vary across task families.

\section{Conclusion}

We asked what information an LLM agent actually needs to carry forward during
interaction, rather than treating its full history as state by default. We
formulated agent state as a compact abstraction of history that preserves the
competence of a fixed reference agent, and instantiated this abstraction as a
deterministic recursive program learned through black-box search. Across
ALFWorld and ScienceWorld, the resulting states often preserve reference-level
task performance while substantially reducing mean policy input tokens, although
the achievable reduction varies considerably across task families. These results
suggest that agent state is neither uniformly small nor identical across tasks,
but can often be represented by a much more selective recurrent summary than the
raw interaction history.

\section*{AI use statement}

Generative AI tools were used to assist with writing and editing the manuscript,
including improving clarity, readability, and organization. They were also used
to assist with software development, including implementation of the genetic
search framework for learning state-abstraction functions and adapting baseline
methods, such as ACON, to environments not supported by their original
implementations.

Generative language models are also an explicit component of the methodology
and experimental setup in this work. We evaluate LLM-based agents, and our
search procedure uses LLMs as proposer and reflector models to generate and
revise candidate state-update programs. The ACON baseline similarly uses
language models as part of its compression and optimization procedure, as
described in the paper.

All AI-assisted manuscript text, code, experimental configurations, and
implementations were reviewed by the authors. The authors verified the
correctness of the final implementations and experimental results and take
responsibility for all content, claims, analyses, and artifacts presented in
this work.

\subsubsection*{Acknowledgments}
We thank Jayakumar Subramanian (Adobe Research) for his input on formalizing the state abstraction objective and for his guidance throughout this work.

\bibliography{iclr2027_conference,brano}

\begin{thebibliography}{35}
\providecommand{\natexlab}[1]{#1}
\providecommand{\url}[1]{\texttt{#1}}
\expandafter\ifx\csname urlstyle\endcsname\relax
  \providecommand{\doi}[1]{doi: #1}\else
  \providecommand{\doi}{doi: \begingroup \urlstyle{rm}\Url}\fi

\bibitem[Abel et~al.(2016)Abel, Hershkowitz, and Littman]{abel2016near}
David Abel, David Hershkowitz, and Michael Littman.
\newblock Near optimal behavior via approximate state abstraction.
\newblock In \emph{International Conference on Machine Learning}, pp.\
  2915--2923. PMLR, 2016.

\bibitem[Agrawal et~al.(2026)Agrawal, Tan, Soylu, Ziems, Khare, Opsahl-Ong,
  Singhvi, Shandilya, Ryan, Jiang, et~al.]{agrawal2026gepa}
Lakshya~A Agrawal, Shangyin Tan, Dilara Soylu, Noah Ziems, Rishi Khare, Krista
  Opsahl-Ong, Arnav Singhvi, Herumb Shandilya, Michael~J Ryan, Meng Jiang,
  et~al.
\newblock Gepa: Reflective prompt evolution can outperform reinforcement
  learning.
\newblock In \emph{International Conference on Learning Representations},
  volume 2026, pp.\  8479--8565, 2026.

\bibitem[Bellman(1957)]{bellman57dynamic}
Richard Bellman.
\newblock \emph{Dynamic Programming}.
\newblock Princeton University Press, Princeton, NJ, 1957.

\bibitem[Castro(2020)]{castro2020scalable}
Pablo~Samuel Castro.
\newblock Scalable methods for computing state similarity in deterministic
  markov decision processes.
\newblock In \emph{Proceedings of the AAAI Conference on Artificial
  Intelligence}, volume~34, pp.\  10069--10076, 2020.

\bibitem[Chevalier et~al.(2023)Chevalier, Wettig, Ajith, and
  Chen]{chevalier2023adapting}
Alexis Chevalier, Alexander Wettig, Anirudh Ajith, and Danqi Chen.
\newblock Adapting language models to compress contexts.
\newblock In \emph{Proceedings of the 2023 Conference on Empirical Methods in
  Natural Language Processing}, pp.\  3829--3846, 2023.

\bibitem[Chevalier-Boisvert et~al.(2019)Chevalier-Boisvert, Bahdanau, Lahlou,
  Willems, Saharia, Nguyen, and Bengio]{chevalierboisvert2019babyai}
Maxime Chevalier-Boisvert, Dzmitry Bahdanau, Salem Lahlou, Lucas Willems,
  Chitwan Saharia, Thien~Huu Nguyen, and Yoshua Bengio.
\newblock {BabyAI}: A platform to study the sample efficiency of grounded
  language learning.
\newblock In \emph{International Conference on Learning Representations}, 2019.

\bibitem[Ferns et~al.(2012)Ferns, Panangaden, and Precup]{ferns2012metrics}
Norman Ferns, Prakash Panangaden, and Doina Precup.
\newblock Metrics for finite markov decision processes.
\newblock \emph{arXiv preprint arXiv:1207.4114}, 2012.

\bibitem[Guo et~al.(2024)Guo, Wang, Guo, Li, Song, Tan, Liu, Bian, and
  Yang]{guo2024evoprompt}
Qingyan Guo, Rui Wang, Junliang Guo, Bei Li, Kaitao Song, Xu~Tan, Guoqing Liu,
  Jiang Bian, and Yujiu Yang.
\newblock Connecting large language models with evolutionary algorithms yields
  powerful prompt optimizers.
\newblock In \emph{International Conference on Learning Representations
  (ICLR)}, 2024.

\bibitem[Holland(1975)]{holland1975misc-adaptation}
John~H. Holland.
\newblock {Adaptation in Natural and Artificial Systems}.
\newblock University of Michigan Press, 1975.
\newblock URL
  \url{https://mlanthology.org/misc/1975/holland1975misc-adaptation/}.

\bibitem[Jiang et~al.(2023)Jiang, Wu, Lin, Yang, and Qiu]{jiang2023llmlingua}
Huiqiang Jiang, Qianhui Wu, Chin-Yew Lin, Yuqing Yang, and Lili Qiu.
\newblock Llmlingua: Compressing prompts for accelerated inference of large
  language models.
\newblock In \emph{Proceedings of the 2023 conference on empirical methods in
  natural language processing}, pp.\  13358--13376, 2023.

\bibitem[Kang et~al.(2025)Kang, Chen, Han, Inan, Wutschitz, Chen, Sim, and
  Rajmohan]{kang2025acon}
Minki Kang, Wei-Ning Chen, Dongge Han, Huseyin~A Inan, Lukas Wutschitz, Yanzhi
  Chen, Robert Sim, and Saravan Rajmohan.
\newblock Acon: Optimizing context compression for long-horizon llm agents.
\newblock \emph{arXiv preprint arXiv:2510.00615}, 2025.

\bibitem[Li et~al.(2006)Li, Walsh, and Littman]{li2006towards}
Lihong Li, Thomas~J Walsh, and Michael~L Littman.
\newblock Towards a unified theory of state abstraction for mdps.
\newblock \emph{AI\&M}, 1\penalty0 (2):\penalty0 3, 2006.

\bibitem[Liang et~al.(2026)Liang, Cao, Zhao, Teng, Liao, Zhao, and
  Liu]{liang2026learningremembermetacognitivemanagement}
Sirui Liang, Pengfei Cao, Jian Zhao, Wenhao Teng, Xiangwen Liao, Jun Zhao, and
  Kang Liu.
\newblock Learning how to remember: A meta-cognitive management method for
  structured and transferable agent memory.
\newblock In \emph{Findings of the Association for Computational Linguistics:
  ACL 2026}, pp.\  30733--30753, 2026.
\newblock URL \url{https://aclanthology.org/2026.findings-acl.1535/}.

\bibitem[Littman \& Sutton(2001)Littman and Sutton]{littman2001predictive}
Michael Littman and Richard~S Sutton.
\newblock Predictive representations of state.
\newblock \emph{Advances in neural information processing systems}, 14, 2001.

\bibitem[{OpenAI}(2026)]{openai2026gpt56luna}
{OpenAI}.
\newblock {GPT-5.6 Luna}.
\newblock \url{https://developers.openai.com/api/docs/models/gpt-5.6-luna},
  2026.
\newblock Accessed: 2026-09-24.

\bibitem[Packer et~al.(2023)Packer, Wooders, Lin, Fang, Patil, Stoica, and
  Gonzalez]{packer2023memgpt}
Charles Packer, Sarah Wooders, Kevin Lin, Vivian Fang, Shishir~G Patil, Ion
  Stoica, and Joseph~E Gonzalez.
\newblock Memgpt: Towards llms as operating systems.
\newblock \emph{arXiv preprint arXiv:2310.08560}, 2023.

\bibitem[Park et~al.(2023)Park, O'Brien, Cai, Morris, Liang, and
  Bernstein]{park2023generative}
Joon~Sung Park, Joseph O'Brien, Carrie~Jun Cai, Meredith~Ringel Morris, Percy
  Liang, and Michael~S Bernstein.
\newblock Generative agents: Interactive simulacra of human behavior.
\newblock In \emph{Proceedings of the 36th annual acm symposium on user
  interface software and technology}, pp.\  1--22, 2023.

\bibitem[Prasad et~al.(2024)Prasad, Koller, Hartmann, Clark, Sabharwal, Bansal,
  and Khot]{prasad2024adapt}
Archiki Prasad, Alexander Koller, Mareike Hartmann, Peter Clark, Ashish
  Sabharwal, Mohit Bansal, and Tushar Khot.
\newblock Adapt: As-needed decomposition and planning with language models.
\newblock In \emph{Findings of the Association for Computational Linguistics:
  NAACL 2024}, pp.\  4226--4252, 2024.

\bibitem[{Qwen Team}(2026)]{qwen2026qwen35}
{Qwen Team}.
\newblock Qwen3.5: Towards native multimodal agents.
\newblock
  \url{https://www.alibabacloud.com/blog/qwen3-5-towards-native-multimodal-agents_602894},
  2026.
\newblock Accessed: 2026-11-15.

\bibitem[Romera-Paredes et~al.(2024)Romera-Paredes, Barekatain, Novikov, Balog,
  Kumar, Dupont, Ruiz, Ellenberg, Wang, Fawzi, Kohli, and
  Fawzi]{romeraparedes2024funsearch}
Bernardino Romera-Paredes, Mohammadamin Barekatain, Alexander Novikov, Matej
  Balog, M.~Pawan Kumar, Emilien Dupont, Francisco J.~R. Ruiz, Jordan~S.
  Ellenberg, Pengming Wang, Omar Fawzi, Pushmeet Kohli, and Alhussein Fawzi.
\newblock Mathematical discoveries from program search with large language
  models.
\newblock \emph{Nature}, 625:\penalty0 468--475, 2024.

\bibitem[Rozanov \& Rei(2025)Rozanov and
  Rei]{rozanov2025stateactenhancingllmbase}
Nikolai Rozanov and Marek Rei.
\newblock Stateact: Enhancing llm base agents via self-prompting and
  state-tracking.
\newblock In \emph{Proceedings of the 1st Workshop for Research on Agent
  Language Models (REALM 2025)}, pp.\  367--385, 2025.

\bibitem[Shinn et~al.(2023)Shinn, Cassano, Berman, Gopinath, Narasimhan, and
  Yao]{shinn2023reflexion}
Noah Shinn, Federico Cassano, Edward Berman, Ashwin Gopinath, Karthik
  Narasimhan, and Shunyu Yao.
\newblock Reflexion: Language agents with verbal reinforcement learning.
\newblock In \emph{Advances in Neural Information Processing Systems
  (NeurIPS)}, 2023.

\bibitem[Shridhar et~al.(2021)Shridhar, Yuan, C{\^o}t{\'e}, Bisk, Trischler,
  and Hausknecht]{shridhar2021alfworld}
Mohit Shridhar, Xingdi Yuan, Marc-Alexandre C{\^o}t{\'e}, Yonatan Bisk, Adam
  Trischler, and Matthew Hausknecht.
\newblock {ALFW}orld: Aligning text and embodied environments for interactive
  learning.
\newblock In \emph{International Conference on Learning Representations
  (ICLR)}, 2021.

\bibitem[Singh et~al.(2003)Singh, Littman, Jong, Pardoe, and
  Stone]{singh2003learning}
Satinder~P Singh, Michael~L Littman, Nicholas~K Jong, David Pardoe, and Peter
  Stone.
\newblock Learning predictive state representations.
\newblock In \emph{Proceedings of the 20th International Conference on Machine
  Learning (ICML-03)}, pp.\  712--719, 2003.

\bibitem[Sondik(1971)]{sondik71thesis}
Edward Sondik.
\newblock \emph{The Optimal Control of Partially Observable {Markov} Decision
  Processes}.
\newblock PhD thesis, Stanford University, 1971.

\bibitem[Subramanian et~al.(2022)Subramanian, Sinha, Seraj, and
  Mahajan]{subramanian2022approximate}
Jayakumar Subramanian, Amit Sinha, Raihan Seraj, and Aditya Mahajan.
\newblock Approximate information state for approximate planning and
  reinforcement learning in partially observed systems.
\newblock \emph{Journal of Machine Learning Research}, 23\penalty0
  (12):\penalty0 1--83, 2022.

\bibitem[Sutton(1988)]{sutton88learning}
Richard Sutton.
\newblock Learning to predict by the methods of temporal differences.
\newblock \emph{Machine Learning}, 3:\penalty0 9--44, 1988.

\bibitem[Wang et~al.(2023)Wang, Xie, Jiang, Mandlekar, Xiao, Zhu, Fan, and
  Anandkumar]{wang2023voyager}
Guanzhi Wang, Yuqi Xie, Yunfan Jiang, Ajay Mandlekar, Chaowei Xiao, Yuke Zhu,
  Linxi Fan, and Anima Anandkumar.
\newblock Voyager: An open-ended embodied agent with large language models.
\newblock \emph{arXiv preprint arXiv:2305.16291}, 2023.

\bibitem[Wang et~al.(2022)Wang, Jansen, C{\^o}t{\'e}, and
  Ammanabrolu]{wang2022scienceworld}
Ruoyao Wang, Peter Jansen, Marc-Alexandre C{\^o}t{\'e}, and Prithviraj
  Ammanabrolu.
\newblock Scienceworld: Is your agent smarter than a 5th grader?
\newblock In \emph{Proceedings of the 2022 Conference on Empirical Methods in
  Natural Language Processing}, pp.\  11279--11298, 2022.

\bibitem[Watkins \& Dayan(1992)Watkins and Dayan]{watkins92qlearning}
Christopher Watkins and Peter Dayan.
\newblock Q-learning.
\newblock \emph{Machine Learning}, 8\penalty0 (3):\penalty0 279--292, 1992.

\bibitem[Williams(1992)]{williams92simple}
Ronald Williams.
\newblock Simple statistical gradient-following algorithms for connectionist
  reinforcement learning.
\newblock \emph{Machine Learning}, 8\penalty0 (3-4):\penalty0 229--256, 1992.

\bibitem[Wu et~al.(2021)Wu, Ouyang, Ziegler, Stiennon, Lowe, Leike, and
  Christiano]{wu2021recursively}
Jeff Wu, Long Ouyang, Daniel~M Ziegler, Nisan Stiennon, Ryan Lowe, Jan Leike,
  and Paul Christiano.
\newblock Recursively summarizing books with human feedback.
\newblock \emph{arXiv preprint arXiv:2109.10862}, 2021.

\bibitem[Xiong et~al.(2026)Xiong, Hu, and
  Clune]{xiong2026learningcontinuallylearnmetalearning}
Yiming Xiong, Shengran Hu, and Jeff Clune.
\newblock Learning to continually learn via meta-learning agentic memory
  designs, 2026.
\newblock URL \url{https://arxiv.org/abs/2602.07755}.

\bibitem[Yao et~al.(2023)Yao, Zhao, Yu, Du, Shafran, Narasimhan, and
  Cao]{yao2023react}
Shunyu Yao, Jeffrey Zhao, Dian Yu, Nan Du, Izhak Shafran, Karthik Narasimhan,
  and Yuan Cao.
\newblock {ReAct}: Synergizing reasoning and acting in language models.
\newblock In \emph{International Conference on Learning Representations
  (ICLR)}, 2023.

\bibitem[Yuksekgonul et~al.(2024)Yuksekgonul, Bianchi, Boen, Liu, Huang,
  Guestrin, and Zou]{yuksekgonul2024textgrad}
Mert Yuksekgonul, Federico Bianchi, Joseph Boen, Sheng Liu, Zhi Huang, Carlos
  Guestrin, and James Zou.
\newblock Textgrad: Automatic" differentiation" via text.
\newblock \emph{arXiv preprint arXiv:2406.07496}, 2024.

\end{thebibliography}
\bibliographystyle{iclr2027_conference}

\newpage
\appendix
\section{Learned State-Update Programs}
\label{app:phi_examples}

This appendix provides representative excerpts from the learned state-update programs for each task family. We omit normalization, defensive copying, and other implementation boilerplate where possible, and instead show the portions of each program that best illustrate how raw interaction text is converted into persistent state.

\begin{table}[t]
\centering
\small
\setlength{\tabcolsep}{4pt}
\begin{tabular}{p{0.31\linewidth}p{0.61\linewidth}}
\toprule
\textbf{Task family} & \textbf{Persistent information retained} \\
\midrule
\multicolumn{2}{l}{\textit{ALFWorld}} \\
Pick \& Place
& current location, target and destination, held object, search frontier, searched receptacles, task phase \\

Look in Light
& object locations, receptacle contents, open/closed state, device state, examination results, recent observations \\

Clean \& Place
& location, held object, target locations, cleanliness, receptacle contents, searched locations, task completion \\

Heat \& Place
& target identity and location, required condition, object property, container state, task phase \\

Cool \& Place
& target identity and location, required transformation, container state, searched locations, task phase \\

Two Objects
& bounded recent interaction history \\

\midrule
\multicolumn{2}{l}{\textit{ScienceWorld}} \\
Lifespan
& entity identity, location, developmental stage, inspected stage, inventory and task progress \\

Forces
& bounded recent interaction history \\

Classification
& entities, aliases, flower-pot contents, inventory, object locations, facts, milestones \\

Genetics
& object locations, inventory, device states, connections, plant-development stages, failed actions \\
\bottomrule
\end{tabular}
\caption{Representative persistent information retained by the learned state-update programs.}
\label{tab:phi_summary}
\end{table}

\begin{table}[t]
\centering
\small
\setlength{\tabcolsep}{5pt}
\begin{tabular}{lcc}
\toprule
\textbf{Task family} & \textbf{Program LOC} & \textbf{State fields} \\
\midrule
\multicolumn{3}{l}{\textit{ALFWorld}} \\
Pick \& Place       & 278 & 11 \\
Look in Light       & 174 & 11 \\
Clean \& Place      & 297 & 15 \\
Heat \& Place       & 428 & 18 \\
Cool \& Place       & 358 & 18 \\
Two Objects         & 10  & 1  \\
\midrule
\multicolumn{3}{l}{\textit{ScienceWorld}} \\
Lifespan            & 283 & 10 \\
Forces              & 18  & 1  \\
Classification      & 210 & 12 \\
Genetics            & 165 & 15 \\
\bottomrule
\end{tabular}
\caption{
Structural complexity of the final learned state-update programs.
Program Lines Of Code (LOC) counts non-empty, non-comment source lines in the final
$\phi$. State fields reports the number of distinct top-level keys that may
appear in the returned state dictionary; individual states may contain fewer
fields because many keys are conditionally omitted.
}
\label{tab:phi_complexity}
\end{table}

\subsection{ALFWorld}

\paragraph{Pick \& Place.}
The learned abstraction maintains a search frontier and converts observations about receptacles into persistent search state. Finding the target changes the task phase and removes exploratory state that is no longer required.

\begin{phibox}{Pick \& Place: Search State}
m = re.search(r"\byou arrive at\s+(.+?)(?:\.|$)", ol)
if m:
    loc = norm(m.group(1))
    focus = loc
    add_receptacle(loc)
    if held is None:
        phase = "search"

m = re.search(
    r"\b(?:on|in)\s+the\s+([a-z][a-z0-9]*(?:\s+\d+)?)"
    r"\s*,?\s*you see\s+(.*?)(?:\.|$)", ol,
)
if m:
    view = norm(m.group(1))
    contents = m.group(2)
    mark_searched(view)

    found = matching_target(contents)
    if found and held is None:
        objloc = view
        phase = "pickup"
        frontier = []

frontier = [
    x for x in dict.fromkeys(frontier)
    if x not in searched and not is_destination(x)
]
\end{phibox}

\paragraph{Look at Object in Light.}
Here the abstraction maintains a small world model containing object locations, receptacle contents, container state, and device properties. It explicitly bounds these structures and preferentially retains goal-relevant objects.

\begin{phibox}{Look in Light: Structured World State}
loc = dict(new.get("loc", {}))
cont = {k: list(v) for k, v in new.get("cont", {}).items()}
opn = dict(new.get("open", {}))
prop = dict(new.get("prop", {}))
exam = dict(new.get("exam", {}))

for x in ents(lo):
    if re.search(r"\b" + re.escape(x) + r"\s+is closed\b", lo):
        opn[x] = "closed"
    elif re.search(r"\b" + re.escape(x) + r"\s+is open\b", lo):
        opn[x] = "open"

eff = re.search(
    r"\b(?:turn|switch)\s+(on|off)\s+(?:the\s+)?("
    + ent_re + r")", lo,
)
if eff:
    prop[eff.group(2)] = eff.group(1)

if len(loc) > 28:
    keep = [
        k for k in loc
        if k == held or
        (goal_obj and k.startswith(goal_obj + " "))
    ]
    keep += [k for k in loc if k not in keep]
    loc = {k: loc[k] for k in keep[:28]}
\end{phibox}

\paragraph{Clean \& Place.}
The cleaning abstraction tracks whether the target object has acquired the required property, alongside its location and the agent's search history. Once the task is complete, it aggressively prunes search-related state.

\begin{phibox}{Clean \& Place: Property and Completion State}
m = re.search(r"\bclean\s+([a-z]+\s+\d+)\b", am)
if m and not failure:
    obj = m.group(1).lower()
    if "you clean" in low_obs or "cleaned" in low_obs:
        out["clean"][obj] = True

m = re.search(
    r"\b(?:move|put|place)\s+([a-z]+\s+\d+)\s+"
    r"(?:to|in|on)\s+([a-z]+\s+\d+)", am,
)
if m and not failure:
    obj, dst = m.group(1).lower(), m.group(2).lower()
    out["held"] = None
    out["items"][obj] = dst

    if (
        move_confirmed
        and is_target
        and is_destination
        and (not requires_clean or is_clean)
    ):
        out["done"] = True

if out.get("done"):
    out["searched"] = [
        p for p in out["searched"]
        if p in keep_places
    ]
\end{phibox}

\paragraph{Heat \& Place.}
The learned program first parses the task specification to determine the target object and required condition, and then converts successful transformation events into persistent object properties.

\begin{phibox}{Heat \& Place: Goal and Property Parsing}
m = re.search(
    r"\b(?:heat|warm)\s+(?:some|a|an)\s+([a-z]+)",
    gt,
)
if m:
    target, cond = m.group(1), "hot"

effects = (
    ("hot",
     r"\byou heat the ([a-z]+\s+\d+)\b",
     r"\bheat\s+([a-z]+\s+\d+)\s+with\b"),
    ("cold",
     r"\byou (?:cool|chill) the ([a-z]+\s+\d+)\b",
     r"\b(?:cool|chill)\s+([a-z]+\s+\d+)\s+with\b"),
)

for value, op, ap in effects:
    m = re.search(op, lo)
    if not m and not failed():
        m = re.search(ap, la)

    if m and is_target(m.group(1)):
        oid = norm(m.group(1))
        out["obj"] = oid
        out["prop"][oid] = value
        out["phase"] = "conditioned"

out["prop"] = {
    k: v for k, v in out["prop"].items()
    if is_target(k)
}
\end{phibox}

\paragraph{Cool \& Place.}
The cooling abstraction parses both the target transformation and destination from the task specification. As execution progresses, it transitions among search, acquisition, transformation, and placement phases and prunes exploratory state after the target is acquired.

\begin{phibox}{Cool \& Place: Phase-Dependent State}
if not need:
    for w, n in (
        ("cool", "cool"), ("chill", "cool"),
        ("clean", "clean"), ("wash", "clean"),
        ("heat", "heat"), ("warm", "heat"),
    ):
        if re.search(r"\b" + w + r"\b", g):
            need = n
            break

if target_id:
    target_loc = seen.get(
        target_id, out.get("target_loc", "")
    )
    status = statuses.get(
        target_id, out.get("target_status", "")
    )

    if target_loc == "held" or held == target_id:
        if need and status != need:
            phase = "transform"
        elif dest:
            phase = "place"
        else:
            phase = "done"
    elif target_loc:
        phase = "acquire"
    else:
        phase = "search"

if acquired or phase == "done":
    out.pop("target_candidates", None)
    out.pop("searched", None)
    out.pop("source_map", None)
\end{phibox}

\paragraph{Two Objects.}
In contrast to the preceding task families, the selected abstraction remains close to recent interaction history. This is consistent with the much smaller token reduction observed for this task family.

\begin{phibox}{Two Objects: Bounded Recent History}
new_state = dict(state)
history = list(state.get("history", []))

entry = {
    "a": event.get("last_action") or "",
    "o": event.get("last_observation") or "",
}

history.append(entry)
new_state["history"] = history[-40:]

return new_state
\end{phibox}

\subsection{ScienceWorld}

\paragraph{Lifespan.}
The learned lifespan abstraction canonicalizes entities across textual descriptions and maintains their current developmental stage, location, and whether the relevant stage has been explicitly inspected.

\begin{phibox}{Lifespan: Entity and Developmental State}
def remember_entity(surface, inspected=False):
    surface = clean(surface).lower()
    key = canonical(surface)
    stage = lifecycle_stage(surface)

    old_entity = out["entities"].get(key, {})
    previous_stage = old_entity.get("s", "")

    record = {
        "s": stage,
        "n": surface,
        "l": out["loc"] or old_entity.get("l", "unknown"),
        "is": old_entity.get("is", ""),
    }

    if previous_stage and previous_stage != stage:
        record["is"] = ""

    if inspected:
        record["is"] = stage

    out["entities"][key] = record
\end{phibox}

\paragraph{Forces.}
As with ALFWorld two-object placement, the selected abstraction for forces remains a bounded interaction history rather than reducing the trajectory to a small structured state.

\begin{phibox}{Forces: Bounded Recent History}
new_state = dict(state)
history = list(state.get("history", []))

entry = {
    "a": event.get("last_action") or "",
    "o": event.get("last_observation") or "",
}

history.append(entry)
new_state["history"] = history[-40:]

return new_state
\end{phibox}

\paragraph{Classification.}
The classification abstraction constructs a compact entity-relation representation. It parses flower-pot contents, resolves generic references to concrete entities, tracks locations and inventory, and bounds the number of retained facts and milestones.

\begin{phibox}{Classification: Entity Resolution}
for match in re.finditer(
    r"flower pot\s+(\d+)\s*\(containing\s+([^)]*)\)",
    obs, flags=re.I
):
    pot = "pot" + match.group(1)
    contents = clean(match.group(2))
    new["pots"][pot] = contents

    if contents and contents.lower() not in {
        "nothing", "empty"
    }:
        entity = resolve(contents)
        new["ents"][entity.lower()] = {"loc": pot}

if (
    action_obj
    and generic(action_obj)
    and not generic(observed_obj)
):
    new["aliases"][action_obj.lower()] = observed_obj

new["facts"] = new["facts"][-6:]
new["milestones"] = new["milestones"][-10:]
\end{phibox}

\paragraph{Genetics.}
The genetics abstraction retains relational and process state, including connections between objects, device configurations, object locations, and plant-development stages.

\begin{phibox}{Genetics: Relational and Process State}
m = re.search(r"connect (.+?) to (.+)", action, re.I)
if m and (
    "connect" in ol or
    "connected" in ol or
    "now" in ol
):
    edge = (
        clean(m.group(1), 55).lower()
        + ">"
        + clean(m.group(2), 55).lower()
    )
    if edge not in con:
        con.append(edge)

pot_match = re.search(
    r"(flower pot\s*\d+)", obs, re.I
)
if pot_match:
    pot = clean(pot_match.group(1), 30).lower()

    if "reproducing stage" in ol:
        stage = "reproducing"
    elif "flowering stage" in ol:
        stage = "flowering"
    elif "seedling" in ol:
        stage = "seedling"
    elif "dead" in ol:
        stage = "dead"
    else:
        stage = None

    if stage:
        plants[pot] = stage
\end{phibox}

\paragraph{Summary.}
Across task families, the learned programs repeatedly convert raw observations into persistent structured variables and remove textual evidence once its relevant information has been incorporated. The particular representation is strongly task dependent: ALFWorld abstractions primarily recover navigation, search, object, and transformation state, while ScienceWorld additionally requires entity resolution, relational structure, device configuration, and process variables. For the two task families with little or no context reduction, the selected representations instead remain close to bounded interaction histories.
\section{Additional Search Details}
\label{app:search}

\subsection{Genetic Search Procedure}
\label{app:ga_search}

Algorithm~\ref{alg:ga_search} gives the complete genetic search procedure described in Section~\ref{sec:program_search}.

\begin{algorithm}[t]
\caption{Genetic search for recursive state abstractions}
\label{alg:ga_search}
\begin{algorithmic}[1]
\REQUIRE Initial population $\mathcal{P}_0$, fixed policy $\pi$,
naive full-history reference agent, generations $G$,
maximum allowed success drop $\epsilon$

\STATE Obtain and cache reference rollouts
$\mathcal{D}_{\mathrm{ref}}$

\FORALL{$\phi \in \mathcal{P}_0$}
    \STATE Evaluate $\phi$ once and cache its rollouts
    $\mathcal{D}_{\phi}$ and fitness $F(\phi)$
\ENDFOR

\FOR{$g = 1,\ldots,G$}
    \STATE Select genetic operator
    $o \in \{\textsc{Mutation},\textsc{Crossover}\}$

    \IF{$o = \textsc{Mutation}$}
        \STATE Select parent $\phi_p$ from $\mathcal{P}_{g-1}$
        using tournament selection
        \STATE Retrieve cached rollout $\mathcal{D}_{\phi_p}$
        \STATE Generate reflector feedback using
        $\mathcal{D}_{\phi_p}$ and $\mathcal{D}_{\mathrm{ref}}$
        \STATE $\phi^{(0)} \leftarrow
        \textsc{Mutate}\!\left(
            \phi_p,
            \mathcal{D}_{\phi_p},
            \mathcal{D}_{\mathrm{ref}},
            \text{feedback}
        \right)$
    \ELSE
        \STATE Select parents $\phi_p,\phi_q$ from $\mathcal{P}_{g-1}$
        using tournament selection
        \STATE Retrieve cached rollouts
        $\mathcal{D}_{\phi_p}, \mathcal{D}_{\phi_q}$
        \STATE Generate reflector feedback using
        $\mathcal{D}_{\phi_p}$, $\mathcal{D}_{\phi_q}$,
        and $\mathcal{D}_{\mathrm{ref}}$
        \STATE $\phi^{(0)} \leftarrow
        \textsc{Crossover}\!\left(
            \phi_p,\phi_q,
            \mathcal{D}_{\phi_p},
            \mathcal{D}_{\phi_q},
            \mathcal{D}_{\mathrm{ref}},
            \text{feedback}
        \right)$
    \ENDIF

    \STATE Evaluate $\phi^{(0)}$ once and cache
    $\mathcal{D}_{\phi^{(0)}}$ and $F(\phi^{(0)})$
    \STATE $\mathcal{A} \leftarrow \{\phi^{(0)}\}$

    \FOR{$r = 0,\ldots,2$}
        \IF{$\phi^{(r)}$ satisfies Eq.~\ref{eq:search_feasibility}}
            \STATE \textbf{break}
        \ENDIF

        \STATE Generate reflector feedback using cached
        $\mathcal{D}_{\phi^{(r)}}$, parent rollout(s),
        and $\mathcal{D}_{\mathrm{ref}}$
        \STATE $\phi^{(r+1)} \leftarrow
        \textsc{Repair}(\phi^{(r)},\text{feedback})$
        \STATE Evaluate $\phi^{(r+1)}$ once and cache
        $\mathcal{D}_{\phi^{(r+1)}}$ and $F(\phi^{(r+1)})$
        \STATE $\mathcal{A} \leftarrow
        \mathcal{A} \cup \{\phi^{(r+1)}\}$
    \ENDFOR

    \STATE $\phi^\star \leftarrow
    \arg\max_{\phi \in \mathcal{A}} F(\phi)$
    \STATE Add only $\phi^\star$ to the population
    \STATE $\mathcal{P}_g \leftarrow
    \textsc{UpdatePopulation}(
        \mathcal{P}_{g-1},\phi^\star
    )$
\ENDFOR

\RETURN $\displaystyle
\arg\max_{\phi \in \cup_{g=0}^{G}\mathcal{P}_g} F(\phi)$
\end{algorithmic}
\end{algorithm}

\subsection{Search Hyperparameters}
\label{app:search_hyperparameters}

Table~\ref{tab:ga_config} reports the configuration used for genetic search.
Unless otherwise stated, the same search procedure is used across task families.
We initialize the population with a single naive aggregation function $\phi$,
use a state-size coefficient of $\lambda=5\times10^{-4}$, and define a candidate
as feasible if its task success is within 10\% of the full-history reference
while using fewer policy input tokens than the naive aggregation baseline.
We run 10 search iterations for ALFWorld and 5 for ScienceWorld.
\begin{table}[t]
\centering
\small
\setlength{\tabcolsep}{5pt}
\begin{tabular}{ll}
\toprule
\textbf{Hyperparameter} & \textbf{Value} \\
\midrule
Initial population size & 1 \\
Population size & 8 \\
Search iterations & 10 (ALFWorld), 5 (ScienceWorld) \\
Tournament size & 3 \\
Crossover probability & 0.5 \\
Proposer model & GPT-5.6 Luna \\
Reflector model & GPT-5.6 Luna \\
Rollouts per candidate & 1 \\
Maximum repair attempts & 3 \\
Maximum success drop $\epsilon$ & 10\% \\
State-size coefficient $\lambda$ & $5\times10^{-4}$ \\
Schema $S$ & Frozen \\
Search seed & 0 \\
\bottomrule
\end{tabular}
\caption{Hyperparameters used for genetic search over state-update programs.}
\label{tab:ga_config}
\end{table}
\section{Prompt Templates}
\label{app:prompts}

Our genetic search uses an LLM reflector and proposer, both instantiated with
GPT-5.6 Luna. In all reported runs, the schema $S$ is frozen and only the
state-update program $\phi$ is modified. The reflector receives the environment
description, optimization objective, paired full-history and compact-state
rollouts, and an instruction to diagnose missing state information or
compression opportunities. The proposer then receives the current $\phi$, the
reflector feedback, example episodes, the frozen schema fields, and the program
contract, and proposes a targeted revision to $\phi$.

\subsection{Reflector Prompt}
\label{app:reflector_prompt}

The reflector receives the environment description, optimization objective,
and paired full-history and compact-state trajectories. Its role is to diagnose
which information is missing from the current abstraction when the policy
fails, and to identify compression opportunities when the policy succeeds.
The prompt used in our experiments is reproduced below, with
\texttt{\{...\}} denoting task- or candidate-specific content.

\begin{quote}\small

\texttt{\{environment\_description\}}

\vspace{1mm}
\textbf{Objective.}

What is being optimized: TASK SUCCESS RATE. A candidate is a pair $(S,\phi)$:
$S$ declares fields and a
\texttt{parse\_step(observation, action)} function that extracts one step's
fields from that observation/action pair alone, while
$\phi(\texttt{state},\texttt{event},S,\texttt{goal})$ folds one new parsed
event into a persistent state carried forward across steps.

The fixed policy reads $\phi$'s compact state instead of the full raw
interaction history available to the teacher reference. Success is the primary
objective, while state size provides secondary pressure.

The teacher's specific trajectory is only one example of a successful path.
Different valid trajectories may solve the same task. Do NOT recommend matching
the teacher's step-by-step actions. Focus on whether the candidate solves the
task and, when it fails, why: what information was missing that the policy
needed to choose a better action. Trajectory-shape matching is not the
objective; task completion is.

\vspace{1mm}
\textbf{Reference and candidate trajectories.}

For each evaluation episode, the prompt contains a full-history teacher
trajectory and the corresponding trajectory produced when the policy uses the
candidate abstraction. Each trajectory is rendered step by step as

\begin{verbatim}
--- {label} ---
step i: action={a} -> observation={o}
\end{verbatim}

with available actions and execution errors included when present.

\vspace{1mm}
\textbf{Reflector instruction.}

Your task: analyze the candidate's mixed performance
(\texttt{\{n\_succ\}/\{n\_total\}} succeeded,
\texttt{\{n\_fail\}/\{n\_total\}} failed).

\textbf{FAILURES (priority --- fix these first):}

For each FAILED episode:
\begin{itemize}
    \item What specific step did the policy fail at and why?
    \item What information did the policy need that $\phi$'s state did not contain?
    \item Was this a $\phi$ tracking failure or a missing capability?
\end{itemize}

\textbf{COMPRESSION OPPORTUNITIES (secondary):}

For the SUCCESSFUL episodes, identify any state fields that are redundant or
oversized and safe to remove without breaking correctness.

Be concrete and specific in both sections.

$S$ is FROZEN. Do not suggest changes to \texttt{parse\_step} or $S$ fields.
Focus only on $\phi$.

Do NOT recommend matching the teacher's specific action sequence.

\end{quote}

When all shown episodes fail, the reflector is instead instructed to diagnose
why the candidate fails across the set. When all episodes succeed, it is
instructed to focus on compression opportunities in $\phi$'s state.

\subsection{Proposer Prompt}
\label{app:proposer_prompt}

For mutation, the proposer receives the current state-update program $\phi$,
the reflector feedback, example trajectories, and the frozen schema $S$. Here, $S$ is a fixed interface schema containing only the current observation
and preceding action; \texttt{parse\_step} simply populates these fields and
does not encode task-specific state knowledge. The
prompt used in our experiments is reproduced below, with
\texttt{\{...\}} denoting task- or candidate-specific content.

\begin{quote}\small

\textbf{Contract.}

Implement a single Python function
\texttt{def phi(state, event, S, goal):}
that folds ONE new parsed event into a persistent state, carried forward
across many calls, one per time step.

\texttt{state} is whatever was returned on the previous call, or an initial
empty dictionary on the first step. \texttt{event} contains the parsed fields
for the current step. \texttt{S} is a dictionary containing
\texttt{S["fields"]}, which maps field names to descriptions. You are not
required to return exactly these keys: you may track additional derived fields
of your own, such as a running count or whether some event has ever occurred,
and you may omit unused fields from $S$.

\texttt{goal} is a plain string describing the task. It may be used to decide
what information is worth tracking, but it does not change within an episode.

Do not mutate the input arguments.

\vspace{1mm}
\texttt{\{environment\_description\}}

\vspace{1mm}
\textbf{Objective.}

What is being optimized: TASK SUCCESS RATE.

The fixed policy reads $\phi$'s compact state instead of the full raw
interaction history available to the teacher reference. Success is the primary
objective, while state size provides secondary pressure.

The teacher's specific trajectory is only one example of a successful path.
Different valid trajectories may solve the same task. Do NOT recommend matching
the teacher's step-by-step actions. Focus on whether the candidate solves the
task and, when it fails, why: what information was missing that the policy
needed to choose a better action. Trajectory-shape matching is not the
objective; task completion is.

\vspace{1mm}
\textbf{Example raw episodes (format grounding):}

\texttt{\{demo\_episodes\}}

\vspace{1mm}
\textbf{S fields (frozen; $\phi$ receives these keys in \texttt{event}):}

\texttt{\{s\_fields\}}

\vspace{1mm}
\textbf{Current $\phi$:}

\texttt{\{phi\_code\}}

\vspace{1mm}
\textbf{Reflection on this pairing's behavior:}

\texttt{\{reflection\_text\}}

\vspace{1mm}
\textbf{Size pressure on $\phi$'s output.}

Keep $\phi$'s returned state SMALL.

If the reflector has identified specific information that was missing and
caused task failure, preserve that information even if it increases state
size. Correctness first, compaction second.

The state returned by $\phi$ is JSON-serialized and fed to the frozen policy
at every step; its per-step token count directly penalizes fitness.

\begin{itemize}
    \item Do NOT store derived or redundant values that can be recomputed from
    primitives.
    \item Do NOT keep accumulating stale entries.
    \item Prune fields whose usefulness has passed.
    \item Do NOT duplicate task information across multiple fields.
    \item Prefer compact keys and values because the state is fed to the policy
    as text.
\end{itemize}

\vspace{1mm}
\textbf{Proposer instruction.}

Propose an improved $\phi$ based on the reflection. Make targeted changes that
directly address the issues identified. Preserve the overall structure and
logic of the current $\phi$ where the reflection does not indicate a problem.
Do not restructure or rewrite parts that are working correctly.

Return your answer as a single Python code block containing just the complete
\texttt{phi} function, with no other text.

\end{quote}

For crossover, the same prompt structure is used with two parent programs
provided instead of one.

\section{Additional Evaluation}\label{app:additional_evals}

Unless otherwise stated, confidence intervals use a paired stratified bootstrap over held-out instances within each task family (10{,}000 resamples, seed 0), with 95\% percentile intervals.

\subsection{Success Uncertainty}\label{app:success_uncertainty}

We evaluate $\phi$, full history and the token-matched window three times on the same 254 held-out ALFWorld and ScienceWorld instances, with the same policy and prompts. Since the policy samples its responses, the repeated runs measure run-to-run variability. For paired comparisons, we average each instance's success over the three runs before bootstrapping. Table~\ref{tab:three_runs} reports the results.

\begin{table}[t]
\centering
\footnotesize
\setlength{\tabcolsep}{4pt}
\begin{tabular}{lccc}
\toprule
 & Success & Paired diff.\ $\phi-$method & Paired diff.\ per run \\
Method & (mean $\pm$ std, 3 runs) & (95\% CI) & (run 1 / 2 / 3) \\
\midrule
$\phi$ & $0.819 \pm 0.007$ & -- & -- \\
Full history & $0.802 \pm 0.010$ & $+1.7$ pp $[-1.8, +5.3]$ & $+1.2$ / $+0.4$ / $+3.5$ \\
Token-matched window & $0.719 \pm 0.018$ & $+10.0$ pp $[+5.9, +13.9]$ & $+10.6$ / $+10.6$ / $+8.7$ \\
\bottomrule
\end{tabular}
\caption{Success over three repeated evaluation runs on all 254 held-out instances.}
\label{tab:three_runs}
\end{table}

We do not detect a systematic difference in success between $\phi$ and full history, and the conclusion holds in every individual run (differences of $+0.4$ to $+3.5$ pp). $\phi$ also has the lowest run-to-run standard deviation of the three methods, at 0.7 pp against 1.0 pp for full history and 1.8 pp for the token-matched window. Per family, its standard deviation is lower than that of full history in 6 of 10 families, equal in 3, and higher in only 1.

Table~\ref{tab:three_runs_family} reports the corresponding per-family results.

\begin{table}[t]
\centering
\small
\begin{tabular}{lcccc}
\toprule
Task family & $n$ & $\phi$ & Full history & Token-matched window \\
\midrule
Pick \& Place & 47 & $0.957 \pm 0.021$ & $0.993 \pm 0.012$ & $0.702 \pm 0.021$ \\
Look in Light & 19 & $1.000 \pm 0.000$ & $0.982 \pm 0.030$ & $0.947 \pm 0.053$ \\
Clean \& Place & 46 & $0.732 \pm 0.013$ & $0.696 \pm 0.058$ & $0.703 \pm 0.045$ \\
Heat \& Place & 27 & $0.852 \pm 0.000$ & $0.778 \pm 0.037$ & $0.605 \pm 0.093$ \\
Cool \& Place & 34 & $0.725 \pm 0.017$ & $0.745 \pm 0.017$ & $0.608 \pm 0.061$ \\
Two Objects & 29 & $0.885 \pm 0.020$ & $0.920 \pm 0.020$ & $0.908 \pm 0.020$ \\
Lifespan & 12 & $0.917 \pm 0.000$ & $0.861 \pm 0.127$ & $0.889 \pm 0.096$ \\
Forces & 12 & $0.694 \pm 0.048$ & $0.639 \pm 0.255$ & $0.694 \pm 0.048$ \\
Classification & 16 & $0.458 \pm 0.036$ & $0.438 \pm 0.108$ & $0.417 \pm 0.095$ \\
Genetics & 12 & $0.861 \pm 0.048$ & $0.694 \pm 0.048$ & $0.861 \pm 0.048$ \\
\bottomrule
\end{tabular}
\caption{Success per task family, mean $\pm$ standard deviation over three repeated evaluation runs.}
\label{tab:three_runs_family}
\end{table}

The three runs also explain the gap between 0.42 and 0.75 on Forces in Table~\ref{tab:scienceworld_results}. On Forces, $\phi$ retains the last 40 steps and episodes are capped at 40 steps, so $\phi$ and full history contain the same interaction content. Across the three runs, full history on Forces scores 0.42, 0.92 and 0.58, so the original gap is consistent with run-to-run variability and should not be interpreted as an improvement from the abstraction.

\subsection{Token-Matched Recent-History Baseline}\label{app:recent_history}

We compare against a token-matched recent-history baseline, choosing $k$ separately for each family so that the last-$k$ context approximately matches $\phi$'s policy-input length. Averaged over the three runs, $\phi$ outperforms the token-matched window by $+10.0$ pp (95\% CI $[+5.9, +13.9]$, $n=254$), and it does so in every individual run ($+8.7$ to $+10.6$ pp). Table~\ref{tab:token_matched_window} reports the families with the largest gains.

\begin{table}[t]
\centering
\small
\begin{tabular}{lcc}
\toprule
Task family & Matched $k$ & $\phi-$last-$k$ success \\
\midrule
Pick \& Place & 2 & +25.5 pp \\
Heat \& Place & 6 & +24.7 pp \\
Cool \& Place & 6 & +11.7 pp \\
\midrule
\textbf{Pooled} & -- & \textbf{+10.0 pp} \\
\bottomrule
\end{tabular}
\caption{Comparison with a token-matched recent-history baseline, averaged over three evaluation runs. For each task family, $k$ is chosen so that the last-$k$ context approximately matches the policy-input length of $\phi$.}
\label{tab:token_matched_window}
\end{table}

Differences are smaller in the remaining families, including those where $\phi$ is itself a bounded history (Two Objects and Forces) or where the token-matched window retains a longer recent history.

\subsection{Generalization to Unseen Environments}\label{app:unseen_envs}

On ALFWorld, which provides rooms seen during training and unseen rooms, $\phi$ matches full history on both splits. Table~\ref{tab:seen_unseen} reports the split-wise differences.

\begin{table}[t]
\centering
\small
\begin{tabular}{lccc}
\toprule
Split & $n$ & $\phi-$full history & 95\% CI \\
\midrule
Seen & 68 & +1.5 pp & $[-4.4,+7.4]$ \\
Unseen & 134 & -1.5 pp & $[-7.5,+4.5]$ \\
\bottomrule
\end{tabular}
\caption{Generalization to seen and unseen ALFWorld rooms. We report the pooled success-rate difference between $\phi$ and full-history conditioning within each split, with 95\% bootstrap confidence intervals.}
\label{tab:seen_unseen}
\end{table}

The learned abstraction achieves 0.81 success on seen rooms and 0.86 on unseen rooms, providing no evidence of overfitting under this split.

\subsection{Cross-Model Transfer on ALFWorld}\label{app:cross_model}

We evaluate cross-model transfer by freezing each $\phi$ learned with GPT-5.6 Luna and using it unchanged with Qwen-3.5-397B \citep{qwen2026qwen35}. On ALFWorld, the pooled difference from Qwen with full history is $-1.5$ pp (95\% CI $[-5.9,+2.5]$, $n=202$). Per-family transfer results are shown in Table~\ref{tab:cross_model_transfer}.

\begin{table}[t]
\centering
\small
\begin{tabular}{lccc}
\toprule
Task & Full@Qwen & $\phi$@Qwen & Token reduction \\
\midrule
Pick \& Place & 1.00 & 0.96 & 68\% \\
Look in Light & 0.95 & 0.84 & 6\% \\
Clean \& Place & 0.83 & 0.78 & 46\% \\
Heat \& Place & 0.78 & 0.85 & 69\% \\
Cool \& Place & 0.94 & 0.94 & 64\% \\
Two Objects & 0.86 & 0.90 & 0\% \\
\bottomrule
\end{tabular}
\caption{Cross-model transfer on ALFWorld, reported separately by task family. Each state-update program $\phi$ is learned with GPT-5.6 Luna and evaluated unchanged with Qwen-3.5-397B. Full@Qwen denotes Qwen with the complete interaction history; $\phi$@Qwen denotes Qwen conditioned on the transferred state abstraction.}
\label{tab:cross_model_transfer}
\end{table}

Cross-model transfer is not guaranteed by our policy-relative notion of sufficiency: a state learned for one policy may omit information used by another. This mismatch may be especially pronounced for a weaker agent that depends more heavily on explicit information a stronger agent can infer. Re-running the search with the new policy remains the intended use when changing agents.
\section{Long-Horizon Evaluation}\label{app:long_horizon}

\subsection{Horizon Scaling on ALFWorld and ScienceWorld}\label{app:horizon_scaling}

To measure how the policy input grows over an episode, we record the number of input tokens in every policy call. At each step $t$, we average this number over all held-out episodes that are still running at step $t$, and drop steps where fewer than three episodes remain. We then fit a least-squares line to these averages as a function of $t$; its slope gives the number of tokens added to the policy input per step.

\begin{figure}[t]
\centering
\begin{tikzpicture}
\begin{axis}[
  xbar, bar width=4pt, width=0.7\textwidth, height=8cm,
  xlabel={Tokens added per step},
  symbolic y coords={Genetics,Classification,Forces,Lifespan,Two Objects,Cool \& Place,Heat \& Place,Clean \& Place,Look in Light,Pick \& Place},
  ytick=data, y tick label style={font=\small},
  enlarge y limits=0.06, xmin=0, xmax=40,
  area legend,
  legend style={at={(0.5,1.02)}, anchor=south, legend columns=2, draw=none},
  legend cell align=left,
]
\addplot[fill=red!60] coordinates {
  (27.3,Pick \& Place) (28.8,Look in Light) (28.0,Clean \& Place) (28.5,Heat \& Place) (28.7,Cool \& Place)
  (27.6,Two Objects) (35.9,Lifespan) (31.4,Forces) (23.8,Classification) (36.4,Genetics)};
\addlegendentry{Full history}
\addplot[fill=blue!60] coordinates {
  (1.8,Pick \& Place) (12.8,Look in Light) (5.6,Clean \& Place) (4.1,Heat \& Place) (4.1,Cool \& Place)
  (18.4,Two Objects) (8.1,Lifespan) (26.2,Forces) (4.4,Classification) (7.2,Genetics)};
\addlegendentry{$\phi$}
\end{axis}
\end{tikzpicture}
\caption{Growth of policy input tokens with episode step on held-out ALFWorld and ScienceWorld episodes. Each bar is the slope of a least-squares fit of the average tokens per call against episode step.}
\label{fig:growth_rates}
\end{figure}
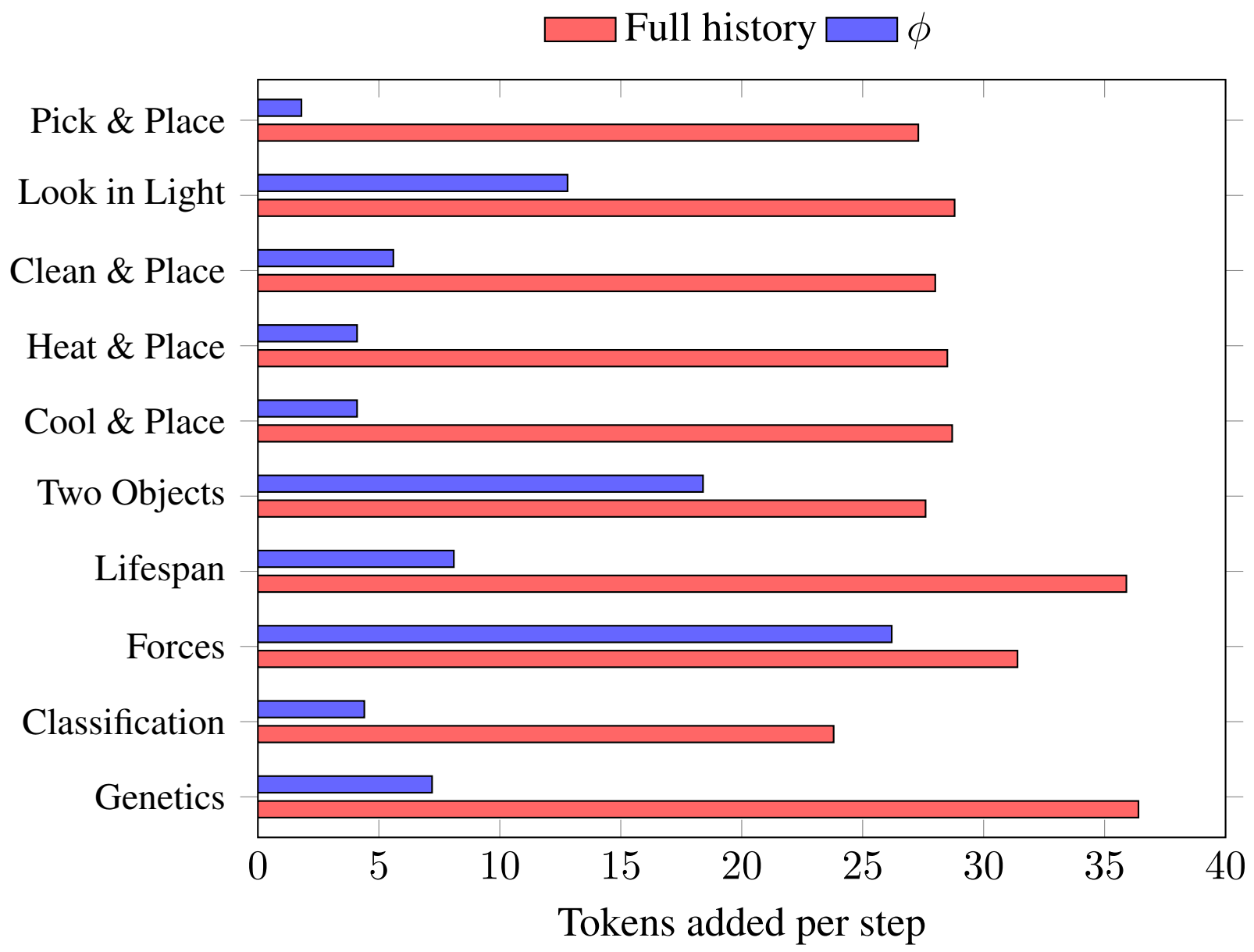

Under full history, the policy input grows by 24--36 tokens per step in every family. Under $\phi$, it grows far more slowly, except in Two Objects and Forces, where $\phi$ is itself a bounded recent history (Figure~\ref{fig:growth_rates}).

\subsection{BabyAI Setup}\label{app:babyai_setup}

We evaluate on BabyAI-GoTo-v0 \citep{chevalierboisvert2019babyai},\footnote{\url{https://minigrid.farama.org/environments/babyai/GoTo}} which has $3\times3$ rooms and a 576-step cap. We use 12 training seeds for search and 50 held-out seeds for evaluation, with the same policy, search settings and naive history aggregator as starting program as in the main experiments.

\subsection{BabyAI Results}\label{app:babyai_results}

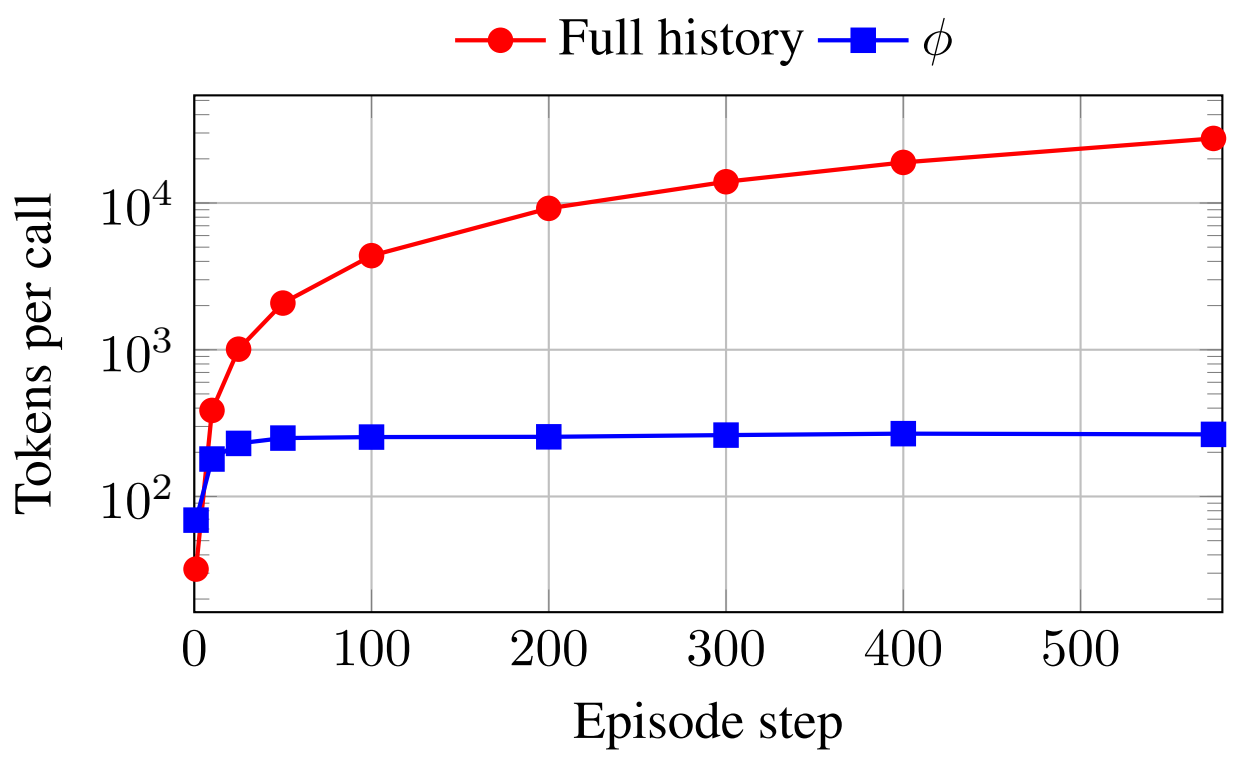
\begin{figure}[t]
\centering
\begin{tikzpicture}
\begin{axis}[
  width=0.6\textwidth, height=5cm,
  xlabel={Episode step}, ylabel={Tokens per call},
  ymode=log, xmin=0, xmax=580,
  legend style={at={(0.5,1.02)}, anchor=south, legend columns=2, draw=none},
  legend cell align=left, grid=major,
]
\addplot[thick, mark=*, color=red] coordinates {
  (1,32) (10,386) (25,1009) (50,2080) (100,4377) (200,9194) (300,13961) (400,18906) (575,27547)};
\addlegendentry{Full history}
\addplot[thick, mark=square*, color=blue] coordinates {
  (1,69) (10,180) (25,230) (50,250) (100,254) (200,255) (300,262) (400,268) (575,265)};
\addlegendentry{$\phi$}
\end{axis}
\end{tikzpicture}
\caption{Policy input tokens per call over an episode on BabyAI-GoTo-v0, averaged over held-out episodes still running at each step.}
\label{fig:babyai_tokens}
\end{figure}

Using the same procedure as in App.~\ref{app:horizon_scaling}, full history adds 48.5 tokens per step and reaches 27.5k tokens per call at the step cap. $\phi$ stays at about 250 tokens per call from step 25 onward (Figure~\ref{fig:babyai_tokens}, Table~\ref{tab:babyai}). The savings grow with the horizon: averaged over all policy calls, $\phi$ reduces mean policy input by 95\% relative to full history on BabyAI, compared with up to 71\% on ALFWorld and 57\% on ScienceWorld, whose episodes are much shorter. Because full history grows with every step while $\phi$ stays bounded, the longer the interaction, the larger the advantage of carrying a compact state instead of the raw history. These savings come at no detectable cost in task success: the difference from full history is not statistically significant (exact McNemar test on paired held-out episodes, $p=0.18$).

\begin{table}[t]
\centering
\small
\begin{tabular}{lcc}
\toprule
 & Full history & $\phi$ \\
\midrule
Mean tokens per call & 3{,}967 & 218 (95\%) \\
Tokens per call at step cap & 27.5k & $\approx$250 \\
Tokens added per step & 48.5 & 0.06 \\
Tokens per episode & 1.89M & 66k \\
\bottomrule
\end{tabular}
\caption{Policy input scaling on BabyAI-GoTo-v0. The number in parentheses is the reduction in mean policy input tokens per call relative to full history, as in Tables~\ref{tab:alfworld_results}--\ref{tab:scienceworld_results}. $\phi$ also uses 29$\times$ fewer input tokens per episode.}
\label{tab:babyai}
\end{table}

\subsection{Learned State for BabyAI}\label{app:babyai_phi}

The learned $\phi$ for BabyAI-GoTo-v0 maintains a compact spatial state: it tracks the agent's heading and position by dead reckoning from its actions, and records where the target was last seen, so the target is not lost when it leaves view. It also keeps a short memory of observed doors and recent failed moves. The excerpt below shows the spatial and target memory.

\begin{phibox}{BabyAI GoTo: Spatial and Target Memory}
# Track heading from turn actions (0=N, 1=E, 2=S, 3=W).
if action == "turn left":
    heading = (heading - 1) % 4
elif action == "turn right":
    heading = (heading + 1) % 4

# Dead reckoning: update position when a forward move succeeds.
if action == "go forward" and not blocked:
    if old_heading == 0:
        y += 1
    elif old_heading == 1:
        x += 1
    elif old_heading == 2:
        y -= 1
    else:
        x -= 1

# Remember where the target was seen, so it is not lost out of view.
if target:
    sightings = list(old.get("tg", []))
    for line in obs.splitlines():
        if label(line) != target:
            continue
        gx, gy, distance = local_position(line)
        sightings.append([gx, gy, distance, heading])
\end{phibox}
\section{Search and State Analysis}\label{app:search_state}

\subsection{Search Statistics}\label{app:search_stats}

To assess how much improvement occurs beyond the initial LLM-generated program, we examine the search records of all ten task families. A single LLM-generated program is generally insufficient: the first program generated in each search satisfies the constraints of \eqref{eq:state_abstraction_objective} in only 2 of 10 families. Across the ten searches, the best program found so far improves 26 times. Every improvement originates from a mutation or crossover proposal: 10 improved on the best program as first proposed (5 from mutation and 5 from crossover), and 16 did so only after reflector-guided repair. Both components therefore contribute: the evolutionary operators generate the candidates from which all improvements arise, and repair turns many otherwise non-improving or infeasible proposals into improvements. In several families, including Pick \& Place, Heat \& Place and Cool \& Place, the final program appears only after multiple generations.

\subsection{Search Cost}\label{app:search_cost}

Table~\ref{tab:search_cost} reports the cost of the search for each task family, measured in meta-LLM calls and in environment rollouts used to evaluate candidate programs. Meta-LLM calls comprise the proposer, which generates and repairs programs, and the reflector, which compares the rollouts of a candidate or its parents with full-history reference rollouts and provides feedback on information the candidate abstraction may have lost (Algorithm~1).

\begin{table}[t]
\centering
\small
\begin{tabular}{lccc}
\toprule
Task family & Meta-LLM calls & Meta-LLM tokens & Rollouts \\
\midrule
Pick \& Place & 126 & 2.0M & 768 \\
Look in Light & 74 & 1.2M & 456 \\
Clean \& Place & 148 & 3.8M & 888 \\
Heat \& Place & 139 & 3.4M & 828 \\
Cool \& Place & 146 & 3.6M & 888 \\
Two Objects & 80 & 1.5M & 492 \\
Lifespan & 33 & 0.4M & 408 \\
Forces & 43 & 0.7M & 504 \\
Classification & 36 & 0.7M & 608 \\
Genetics & 35 & 0.5M & 108 \\
\midrule
Total & 860 & 17.7M & 5{,}948 \\
\midrule
BabyAI GoTo & 52 & 8.6M & 324 \\
\bottomrule
\end{tabular}
\caption{Search cost per task family, including the long-horizon BabyAI task (App.~\ref{app:long_horizon}). Rollouts are the environment episodes used to evaluate candidate programs during search.}
\label{tab:search_cost}
\end{table}

\subsection{Field Usage}\label{app:field_usage}

To see which parts of the learned state the policy explicitly refers to, we analyze every policy call in the held-out evaluations. For each call, we check whether the policy's written reasoning or chosen action names something stored in each field of the state. We group fields into categories that are shared across task families, since each family's program names its fields differently. Table~\ref{tab:field_usage} reports, for each category, its share of all field mentions in the domain and the average number of mentions per step. This measures how often the policy uses each kind of field, not whether removing it would change success.

\begin{table}[t]
\centering
\small
\begin{tabular}{lcccc}
\toprule
 & \multicolumn{2}{c}{ALFWorld} & \multicolumn{2}{c}{ScienceWorld} \\
Category & Share & Per step & Share & Per step \\
\midrule
Target / goal & 22.8\% & 0.97 & 10.0\% & 0.23 \\
Current location & 16.3\% & 0.66 & 7.3\% & 0.17 \\
Exploration memory & 20.4\% & 0.72 & 0.0\% & 0.00 \\
Object \& container knowledge & 21.0\% & 0.83 & 42.5\% & 0.85 \\
Inventory & 3.8\% & 0.19 & 4.8\% & 0.06 \\
Last action / result & 2.8\% & 0.10 & 13.4\% & 0.30 \\
Progress \& control & 2.4\% & 0.13 & 11.1\% & 0.17 \\
Raw observation text & 10.5\% & 0.41 & 10.9\% & 0.30 \\
\midrule
All fields & 100\% & 4.02 & 100\% & 2.09 \\
\bottomrule
\end{tabular}
\caption{How often the policy refers to each category of state field, over all held-out policy calls. Raw observation text denotes retained observation snippets whose contents cannot be assigned to a single field category. Two Objects and Forces are omitted, since their $\phi$ is a single bounded-history field.}
\label{tab:field_usage}
\end{table}

Across the structured abstractions, a small set of task-relevant categories accounts for most explicit references. In ALFWorld these are primarily the target, current location, exploration memory, and object and container knowledge; in ScienceWorld, object and container knowledge is dominant, with additional references to task progress and recent action outcomes. Because this analysis only detects fields whose contents the policy names, low usage does not imply that a field is unimportant, as the masking results in Appendix~\ref{app:field_masking} show.

\subsection{Field Masking}\label{app:field_masking}

To test which parts of the state the policy needs, we hide one group of fields at a time from the policy. The learned $\phi$ runs unchanged and updates its full state at every step. Before each policy call, we copy the state, delete one field group from the copy, and give the policy only this copy. The full state is what $\phi$ receives at the next step. Field groups follow the categories of Table~\ref{tab:field_usage}, and we evaluate all eight task families whose $\phi$ is a structured program, on the same held-out instances, policy and prompts as the main evaluation. The reference is the success of the unmasked $\phi$, averaged over the three evaluation runs of Appendix~\ref{app:success_uncertainty}. As a check, running the masking pipeline with no field hidden reproduces the success of the unmasked $\phi$. Table~\ref{tab:field_masking} reports the change in success for each hidden field group.

\begin{table}[t]
\centering
\small
\begin{tabular}{lccc}
\toprule
Hidden field group & Families & Change in success & 95\% CI \\
\midrule
Progress \& control & 7 & $-36.1$ pp & $[-42.3, -29.7]$ \\
Exploration memory & 5 & $-29.1$ pp & $[-35.6, -22.7]$ \\
Recent observations & 4 & $-29.3$ pp & $[-36.6, -22.1]$ \\
Object \& container knowledge & 8 & $-15.5$ pp & $[-20.3, -10.6]$ \\
Last action / result & 5 & $-10.0$ pp & $[-17.1, -2.9]$ \\
Inventory & 7 & $-8.5$ pp & $[-12.8, -4.3]$ \\
Current location & 8 & $-5.2$ pp & $[-9.7, -0.8]$ \\
Target / goal & 8 & $+0.5$ pp & $[-3.3, +4.1]$ \\
\bottomrule
\end{tabular}
\caption{Change in success when one field group is hidden from the policy, pooled over the task families whose $\phi$ contains fields in that group.}
\label{tab:field_masking}
\end{table}

Hiding seven of the eight field groups lowers success, with every corresponding confidence interval excluding zero. The largest drops come from progress tracking, recent observations and exploration memory. Progress tracking and exploration memory hold which sub-goal the agent is on and where it has already searched, information a recent-history window can discard. Even fields the policy never explicitly names can be essential. On Clean \& Place, hiding the flags that record which objects are already clean drops success from 0.73 to 0.04, as the agent finds and takes the object but keeps cleaning it until the step limit. Target / goal is the only group with no measurable effect, consistent with the task description already stating the goal in the policy prompt.
\section{Hierarchy of Sufficiency Criteria}\label{app:proof_hierarchy}

\begin{proof}[Proof of Proposition~\ref{prop:hierarchy}]
The first inequality in \eqref{eq:relaxation_chain} follows from $x \le |x|$ for any real $x$.

For the second inequality, let $r_t$ denote the reward at step $t$, equal to the binary success indicator at the final step and zero otherwise. Let $Q_t(h_t, a)$ be the action-value function of $\pi_{\mathrm{full}}$ at step $t$, and let
\begin{equation*}
V_t(h_t) = \mathbb{E}_{a \sim \pi_{\mathrm{full}}(\cdot \mid h_t)} Q_t(h_t, a).
\end{equation*}
Set $V_{H+1} \equiv 0$. Let $s_t$ be the state that $\phi$ computes from the history $h_t$. Both agents start from the same initial distribution, so $J(\pi_{\mathrm{full}}) = \mathbb{E}_{h_1 \sim d_1^{\pi\circ\phi}}[V_1(h_1)]$. By the Bellman relation $Q_t(h_t, a) = \mathbb{E}\big[r_t + V_{t+1}(h_{t+1}) \mid h_t, a\big]$,
\begin{align*}
J(\pi\circ\phi)
&= \sum_{t=1}^{H} \mathbb{E}_{h_t \sim d_t^{\pi\circ\phi}} \, \mathbb{E}_{a \sim \pi(\cdot \mid s_t)} \big[ r_t \big] \\
&= \sum_{t=1}^{H} \Big( \mathbb{E}_{h_t \sim d_t^{\pi\circ\phi}} \, \mathbb{E}_{a \sim \pi(\cdot \mid s_t)} Q_t(h_t, a) - \mathbb{E}_{h_{t+1} \sim d_{t+1}^{\pi\circ\phi}} \big[ V_{t+1}(h_{t+1}) \big] \Big).
\end{align*}
Subtracting this from $J(\pi_{\mathrm{full}})$ and telescoping the value terms with $V_{H+1} \equiv 0$ gives
\begin{align*}
J(\pi_{\mathrm{full}}) - J(\pi\circ\phi)
&= \sum_{t=1}^{H} \mathbb{E}_{h_t \sim d_t^{\pi\circ\phi}} \Big[ V_t(h_t) - \mathbb{E}_{a \sim \pi(\cdot \mid s_t)} Q_t(h_t, a) \Big] \\
&= \sum_{t=1}^{H} \mathbb{E}_{h_t \sim d_t^{\pi\circ\phi}} \Big[ \mathbb{E}_{a \sim \pi_{\mathrm{full}}(\cdot \mid h_t)} Q_t(h_t, a) - \mathbb{E}_{a \sim \pi(\cdot \mid s_t)} Q_t(h_t, a) \Big].
\end{align*}
With a terminal binary reward, $Q_t(h_t, a) \in [0,1]$ for every $h_t$ and $a$. For any distributions $p$ and $q$ and any function $f$ with values in $[0,1]$, $|\mathbb{E}_p[f] - \mathbb{E}_q[f]| \le \mathrm{TV}(p,q)$. At each history $h_t$, this gives
\begin{equation*}
\Big| \mathbb{E}_{a \sim \pi_{\mathrm{full}}(\cdot \mid h_t)} Q_t(h_t, a) - \mathbb{E}_{a \sim \pi(\cdot \mid s_t)} Q_t(h_t, a) \Big| \le \mathrm{TV}\big( \pi_{\mathrm{full}}(\cdot \mid h_t),\, \pi(\cdot \mid s_t) \big).
\end{equation*}
Substituting this bound and applying the triangle inequality gives
\begin{equation*}
\big| J(\pi_{\mathrm{full}}) - J(\pi\circ\phi) \big| \le \sum_{t=1}^{H} \mathbb{E}_{h_t \sim d_t^{\pi\circ\phi}} \Big[ \mathrm{TV}\big( \pi_{\mathrm{full}}(\cdot \mid h_t),\, \pi(\cdot \mid s_t) \big) \Big] = B(\phi).
\end{equation*}

For the first inclusion, let $\phi \in \mathcal{F}_{\mathrm{beh}}(\delta)$. By \eqref{eq:relaxation_chain}, $|J(\pi_{\mathrm{full}}) - J(\pi\circ\phi)| \le B(\phi) \le \delta$, so $\phi \in \mathcal{F}_{\mathrm{perf}}(\delta)$. For the second, let $\phi \in \mathcal{F}_{\mathrm{perf}}(\delta)$. Then $J(\pi_{\mathrm{full}}) - J(\pi\circ\phi) \le |J(\pi_{\mathrm{full}}) - J(\pi\circ\phi)| \le \delta$, so $\phi \in \mathcal{F}_{\mathrm{signed}}(\delta)$.
\end{proof}

\end{document}